\documentclass[runningheads]{llncs}

\usepackage{amsmath}
\usepackage{amsfonts} 
\usepackage{amsopn}

\DeclareMathOperator{\rank}{rank}
\DeclareMathOperator{\kmax}{kmax}
\usepackage{orcidlink}
\usepackage{caption}
\usepackage{subcaption}
\usepackage{float} 
\usepackage{tabularx} 
\usepackage{booktabs} 
\usepackage{graphicx,epstopdf}
\usepackage[export]{adjustbox}
\usepackage{tikz}
\usetikzlibrary{shapes.geometric, arrows}
\tikzstyle{startstop} = [rectangle, minimum width = 1.5cm, minimum height = 1cm, text centered, draw = black, fill=red!30]
\tikzstyle{between} = [rectangle, rounded corners, minimum width = 1.5cm, minimum height = 1cm, text centered, draw = black, fill=blue!30]
\tikzstyle{arrow} = [thick, ->, >=stealth]
\usepackage{pgfplots}
\pgfplotsset{compat=1.15}
\usepackage{mathrsfs}
\usetikzlibrary{arrows}
\definecolor{qqqqff}{rgb}{0,0,1}

\usepackage[final]{changes}
\colorlet{Changes@Color}{red}
\definechangesauthor[color=teal]{R1} 
\definechangesauthor[color=red]{R2}

\newcommand{\rev}[1]{\added[id=R2]{#1}}
\newcommand{\del}[1]{\deleted[id=R2]{#1}}

\begin{document}

\title{A topological machine learning pipeline for classification}
\titlerunning{A topological machine learning pipeline for classification}

\author{Francesco Conti\inst{1,2}\orcidlink{0000-0001-7393-4267}, Davide Moroni\inst{1}\orcidlink{0000-0002-5175-5126} and Maria Antonietta Pascali\inst{1}\orcidlink{0000-0001-7742-8126}} 

    \authorrunning{F. Conti et al.}
%
\institute{Institute of Information Science and Technologies ``A. Faedo'', \\ National Research Council of Italy,  Pisa (IT)\\ \email{\{Name.Surname\}@isti.cnr.it}
\and Department of Mathematics, University of Pisa,  Pisa (IT)
}

\maketitle              

\begin{abstract}
    In this work we develop a pipeline that associates persistence diagrams to digital data, via the most appropriate filtration for the type of data considered. Using a grid search approach, this pipeline determines optimal representation methods and parameters. The development of such a topological pipeline for machine learning involves two crucial steps, \del{strongly affecting} \rev{that strongly affect} its performance: firstly, digital data must be represented as \rev{an} algebraic object with a proper associated filtration in order to compute its topological summary, the \emph{persistence diagram}. Secondly, the persistence diagram must be transformed with suitable representation methods in order to be introduced in a machine learning algorithm. We assess the performance of our pipeline, and in parallel we compare the different representation methods on popular benchmark datasets. This work is a first step toward\del{s} both an easy and ready-to-use pipeline for data classification using persistent homology and machine learning, and to understand the theoretical reasons why, given a dataset and a task to be performed, a pair (filtration, topological representation) is better than another.

\end{abstract}

\section{Introduction}\label{sec1}
In the last decade, scientific and industrial research \del{have} \rev{has} introduced the need to manage huge quantities of digital data. Numerous studies have therefore arisen to provide appropriate tools for such research\del{es}. In the mathematical field, \rev{T}opological \rev{D}ata \rev{A}nalysis (TDA) has started to play a major role since it provides qualitative and quantitative features to describe the data space from a geometrical point of view. Topological data analysis has its roots in \rev{P}ersistent \rev{H}omology (PH), which combines algebraic topology with discrete Morse theory. Informally, algebraic topology studies the global shape of a space by means of features that are invariant under continuous deformation. These features are essentially the number of $k$-dimensional holes in the space. Persistent homology\rev{,} on the other hand\rev{,} studies the evolution of the data space at different scales of resolution and tracks the topological invariants ($k$-dimensional holes) that form and vanish during this process. These topological invariants encode the global geometrical properties of the data space. The collection of such information is called a \rev{P}ersistent \rev{D}iagram (PD). The idea of injecting geometric information into \rev{M}achine \rev{L}earning (ML) and deep learning is currently a very active field in the scientific community \cite{bronstein2017geometric,monti2017geometric,krizhevsky2012imagenet,bergomi2019towards,conti2022construction}. Suffice it to say that convolutional neural networks are part of this research area. The results provided by TDA have proved very promising \cite{carlsson2009topology,article,bergomi2019towards,tauzin2021giotto,nielson2015topological}. However, the synergy between TDA and ML is quite novel, since the main concept of TDA, namely persistence diagrams, could not be introduced in a\rev{n} ML approach from the get-go. A wide range of transformations have been devised to exploit the capabilities of PDs in ML algorithms \cite{chazal2014stochastic,bubenik2015statistical,umeda2017time,adams2017persistence,chen2019topological,pun2018persistent,corbet2019kernel}. Each of them has been devised with specific requirements. To the best of our knowledge, however, their relative effectiveness in relation to different heterogeneous types of dataset\rev{s} has  not been the subject of extensive study. More importantly, the link between \rev{the} type of data and optimal representation has not yet been addressed. 
\rev{In other words, even if it has been shown in the rich literature on applications of TDA and machine learning that topological features are feasible for the classification purposes of several types of data, it is still not clear which is the best way to exploit the topological descriptive power, when approaching a new classification task.}
In addition, a key element of persistent homology is the choice of a filtration that associates a PD with the data. The \del{first} goal of this work is to investigate the theoretical reasons behind the two main choices that characterize a topological pipeline \rev{in relation to a specific goal}, namely the choice of \del{a} filtration and the choice of \del{a} representation method. \del{In doing so} \rev{Incidentally}, we will develop a topological pipeline with \del{the} different available filtrations and several representation methods with the associated parameters. We will evaluate the results obtained on benchmark datasets, and in parallel compare the accuracy of the different representation methods. \rev{In doing so, we will lay the groundwork for a correlation between filtration - vectorization - task, with particular focus on data type.} We emphasize the fact that the purpose of this research is not in the classification accuracy of these methods \emph{per se}, but rather to provide a basis for understanding why certain representations are better suited for certain types of data.

\section{Mathematical background}\label{sec2}
The core idea of our pipeline has its root in \rev{T}opological \rev{D}ata \rev{A}nalysis (TDA) and \rev{M}achine \rev{L}earning (ML). While ML is an already established and well-known branch of artificial intelligence with a wide variety of applications, TDA is a relatively new field of research which studies the geometrical and topological aspects of data. The aim of this work is to emphasize the value of a topological approach in a machine learning context of data analysis. Therefore a background on machine learning will be assumed and will not be recalled in this section, nor elsewhere in the article. Conversely, this section is mainly aimed at introducing the reader to topological data analysis, and in particular to algebraic topology and persistent homology. Albeit formalities, TDA aims to recognize and analyze patterns within data by means of topology. The main concept of TDA is persistent homology, where the patterns within data are captured across multiple scales. The persistence of a topological feature is the span over different scales of its detectability, and is an indication of its importance. The collection of such persistences is called \rev{the} persistence diagram.

\subsection{Algebraic topology}\label{2subsec1}
\begin{figure}[!ht]
    \centering
     \begin{subfigure}[b]{0.45\textwidth}
     \centering
    \includegraphics[width=0.8\textwidth]{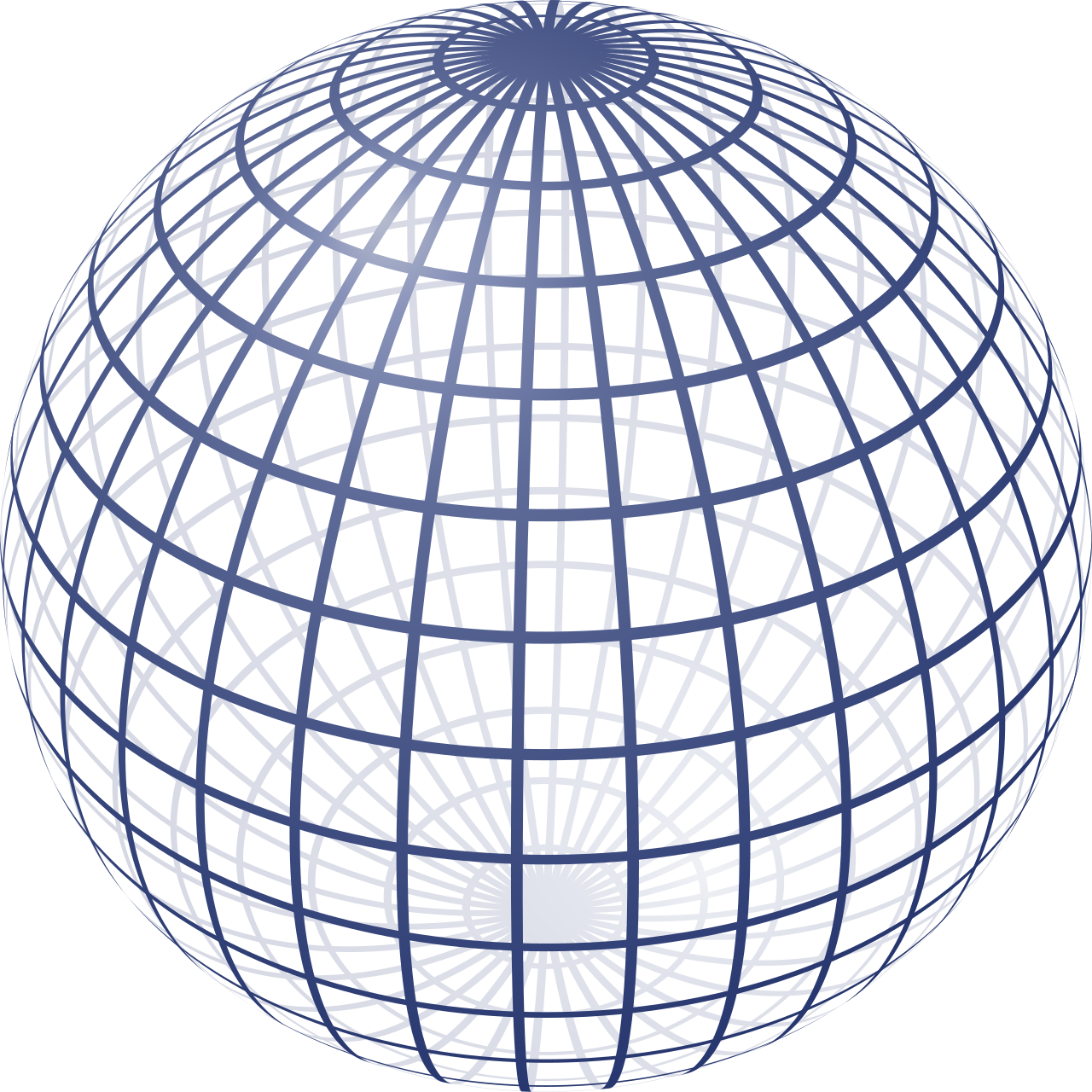}
      \caption{}
      \label{fig:AT_Spehere_Torus:a}
    \end{subfigure}
      \begin{subfigure}[b]{0.45\textwidth}
     \centering
    \includegraphics[width=0.8\textwidth]{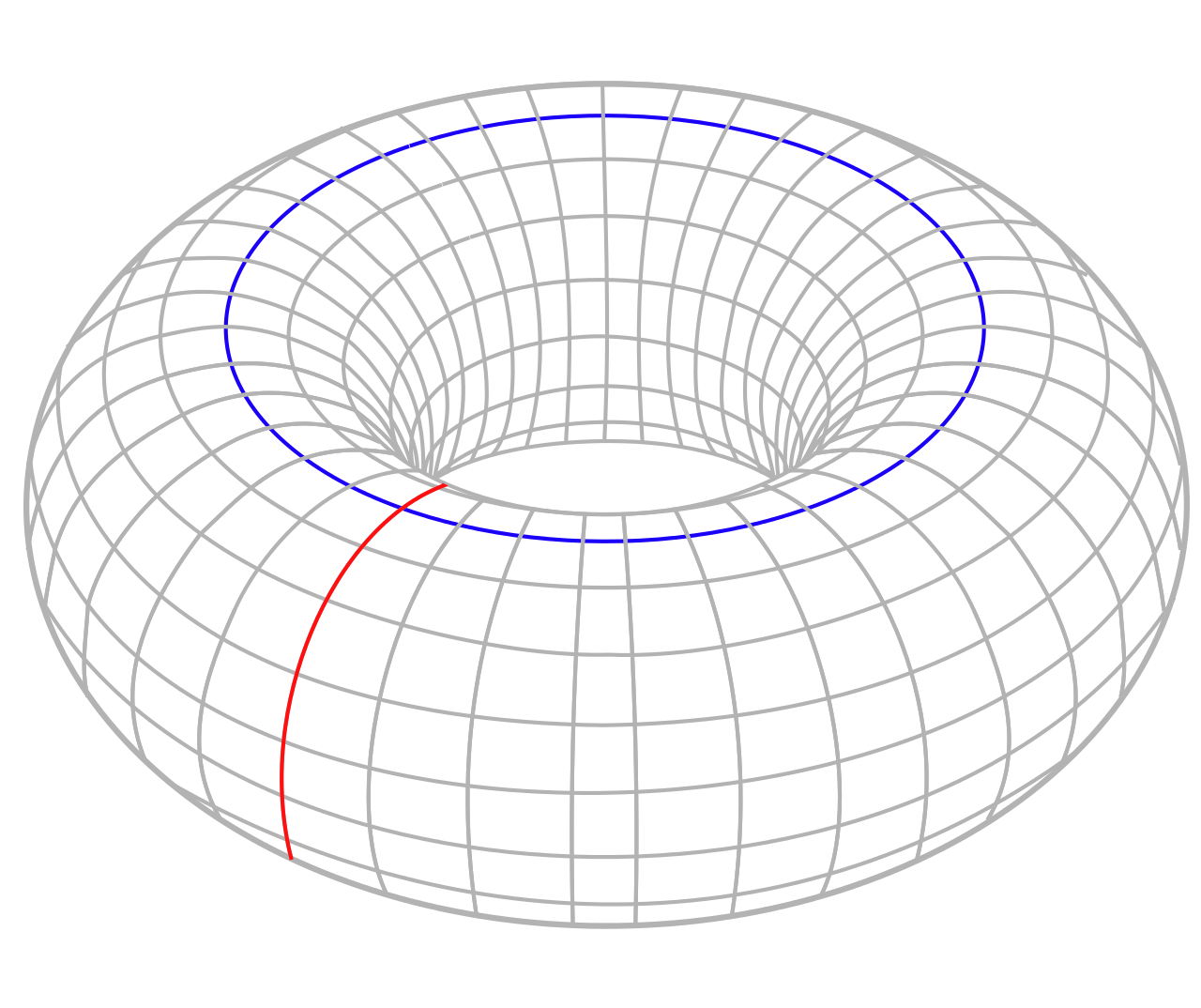}
    \caption{}
    \label{fig:AT_Spehere_Torus:b}
     \end{subfigure}
    \caption{The sphere $S^2$ \rev{(Figure \ref{fig:AT_Spehere_Torus:a})} bounds a $2$-dimensional void. The torus $T^2$ \rev{(Figure \ref{fig:AT_Spehere_Torus:b})} bounds a $2$-dimensional void and two $1$-dimensional holes. Images: Geek3 and YassinMrabet via Wikipedia.}
    \label{fig:AT_Spehere_Torus}
\end{figure}

\del{Albegraic} \rev{Algebraic} topology is the mathematical field which aims to study topological spaces by means of algebraic features that are invariant under continuous deformation. Algebraic topology is a wide mathematical field and only a part of it will be relevant in TDA, therefore this subsection is only a brief introduction to its main concepts. For a more complete guide to algebraic topology, we refer the reader to \cite{MR1867354}. The main concept of algebraic topology that we are going to use is the \emph{homology} of a space, which associates to a topological space a sequence of algebraic objects, namely Abelian groups or modules. More formally, the $k$-th homology group of a topological space $X$ over a field $\mathbb{F}$ is a vector space over $\mathbb{F}$ which we denote as $H_k(X;\mathbb{F})$. Beyond the technicalities of its definition, the rank of $H_k(X;\mathbb{F})$ corresponds to the number of distinct $k$-dimensional holes in the topological space $X$, with the exception of $H_0(X;\mathbb{F})$. In this case, $\rank H_0(X;\mathbb{F})$ corresponds to the number of $0$-dimensional holes plus one, that is, the number of connected components. Other types of homology can be defined, see reduced homology, relative homology\del{,} \rev{and} cohomology. Their definition and use \del{is} \rev{are} beyond the scope of this work, so they will not be addressed. In general, different choices of $\mathbb{F}$ will results in different homology groups $H_k(X;\mathbb{F})$. In this work\rev{,} we limit to the case where $\mathbb{F} = \mathbb{Z}_2$. As an example, \rev{Figure \ref{fig:AT_Spehere_Torus:a}}\del{(left) }shows that the sphere $S^2$ bounds a $2$-dimensional void. $S^2$ is connected and there is no $1$-dimensional hole. The homology group\rev{s} for the sphere are:

    $$H_0(S^2, \mathbb{Z}_2) = \mathbb{Z}  \Rightarrow \rank H_0(S^2, \mathbb{Z}_2) = 1,$$ 
    $$H_1(S^2, \mathbb{Z}_2) = \textbf{0}  \Rightarrow \rank H_1(S^2, \mathbb{Z}_2) = 0,$$ 
    $$H_2(S^2, \mathbb{Z}_2) = \mathbb{Z}  \Rightarrow \rank H_2(S^2, \mathbb{Z}_2) = 1,$$ 
    $$H_k(S^2, \mathbb{Z}_2) = \textbf{0}  \Rightarrow \rank H_k(S^2, \mathbb{Z}_2) = 0 \text{ for every } k \geq 3.$$

As another example, \rev{Figure \ref{fig:AT_Spehere_Torus:b}} \del{(right)} shows that the torus $T^2$ is connected, it bounds two $1$-dimensional holes (the ones inside the blue and the red loop) and one $2$-dimensional hole. The homology group\rev{s} for the torus are:

    $$ H_0(T^2, \mathbb{Z}_2) = \mathbb{Z} \Rightarrow \rank H_0(T^2, \mathbb{Z}_2) = 1, $$
    $$ H_1(T^2, \mathbb{Z}_2) = \mathbb{Z}^2 \Rightarrow \rank H_1(T^2, \mathbb{Z}_2) = 2, $$
    $$ H_2(T^2, \mathbb{Z}_2) = \mathbb{Z} \Rightarrow \rank H_2(T^2, \mathbb{Z}_2) = 1, $$
    $$ H_k(T^2, \mathbb{Z}_2) = \textbf{0} \Rightarrow \rank H_k(T^2, \mathbb{Z}_2) = 0 \text{ for every } k \geq 3.$$

Algebraic topology is a powerful tool that classif\rev{ies} topological spaces with just a finite sequence of integers. The computation for generic topological spaces requires complicated homology and cohomology theories. Persistent homology is now introduced because it enhance\rev{s} the expressiveness of the topological features and\del{, by means of simplicial homology,} it allows easier computation of the homology ranks \rev{by means of \emph{simplicial homology}}.

\subsection{Persistent homology}\label{2subsec2}
Persistent homology studies the geometry of spaces by looking at the evolution of $k$-dimensional holes at different spatial resolutions. The key difference with algebraic topology is that the ranks of the homology groups are computed at various scales and the features we extract are precisely the evolution of such ranks \cite{verri1993use,carlsson2009topology}. This section is only a brief introduction to persistent homology, as many of its concepts will be covered and expanded upon later, when we make actual use in the pipeline. Before continuing, it is necessary to introduce the concept of a simplicial complex. Let $V=\left\{v_0, \dots, v_k\right\}$ be $k+1$ affinely independent points. A $k$-\textit{simplex} $\sigma$ is the convex hull of $V$, which is called the set of vertices of $\sigma$. We call face of $\sigma$ the convex hull of any subset of points of $V$ and we write $\tau \subseteq \sigma$ when $\tau$ is a face of $\sigma$. A \textit{simplicial complex} $\mathcal{K}$ is a set of simplexes such that $\emptyset \in \mathcal{K}$, for every $\sigma \in \mathcal{K}$ and every $\tau \subseteq \sigma$ it holds that  $\tau \in \mathcal{K}$ and the intersection of any two simplexes of $\mathcal{K}$ is always a face of both. A \textit{triangulation} of a topological space $X$ is a couple $(\mathcal{K}, h)$ where $\mathcal{K}$ is a simplicial complex $\mathcal{K}$ and $h\colon \mathcal{K} \to X$ is a homeomorphism. What makes simplicial complexes particularly suitable for our study is that the computation of their homology group\rev{s} is very simple compared to normal topological spaces. In addition, almost all topological spaces found in \del{application} \rev{practice} are triangulable. The ranks of the homology groups of simplicial complexes are more commonly known as Betti numbers. The computation of homology groups and their rank\rev{s} specifically for simplicial complexes is known as \textit{simplicial homology}. For more information on simplicial complexes, simplicial homology and triangulation of topological spaces we refer the reader to \cite{MR1867354}. A \textit{filtration} of a simplicial complex $\mathcal{K}$ is a finite sequence of subcomplexes such that $\emptyset \subseteq \mathcal{K}_0 \subseteq \dots \subseteq \mathcal{K}_n = \mathcal{K}$. Persistent homology keeps track of changes \del{of} \rev{in} the Betti numbers associated to each homology group of $\mathcal{K}_i$ for $i = 0, \dots, n$. The persistence of a topological feature is thus its detectability at different spatial resolutions. In particular, features with high persistence will be important to describe the shape of the data, while those with low persistence will be assimilated to noise. The collection of such features is called \textit{persistence diagram}. To better understand the differences between \rev{p}ersistent homology and algebraic topology, we provide the following example. Suppose we have a normal ball and a ball-shaped sponge, which is of course full of many small holes. The difference between these two objects will be detectable or not depending on the resolution at which we are looking at them. If we are looking at them from a distance, the two objects will be indistinguishable. If we look at them close up, we will see the differences. Moreover, the holes will have limited persistence. Therefore, thanks to persistent homology we can conclude that the two objects have the same shape but one is basically the noisy version of the other. Not only PH provides algebraic topology with easy tools to compute $\rank H_k(\mathcal{K}, \mathbb{F})$ but it also solves the two main problems associated \del{to} \rev{with} digital data, namely their discrete and noisy nature. With regard to the discrete nature, PH allows \del{to associate} \rev{associating} a structure of simplicial complex to discrete data through the filtration. In this way, non-trivial homology groups can be found. As for the noisy nature\rev{,} PH highlights the most persistent structures at different scales, limiting the incidence of noise.

\section{Topological pipeline}\label{sec3}
\begin{figure}
\centering
    \begin{tikzpicture}[node distance = 1.0cm]
    \centering
        \node(Data)[startstop]{Data};
        \node(PDs)[between, right of = Data, xshift = 3cm]{Persistence Diagram};
        \node(Vector)[between, right of = PDs, xshift = 3.5cm]{Vector};
        \node(ML)[startstop, right of = Vector, xshift = 1.0cm]{Classifier};

        \draw [arrow] (Data) -- node[anchor = south] {Filtration} (PDs);
        \draw [arrow] (PDs) -- node[anchor = south] {Vectorization} node[anchor = north] {methods}(Vector);
        \draw [arrow] (Vector) -- (ML);
    \end{tikzpicture}
\caption{The pipeline for a topological study of digital data in a machine learning context. A filtration associates a persistence diagram to the digital data. The persistence diagram is then vectorized by means of various vectorization methods. Finally, the vector is fed to a machine learning classifier.}
\label{fig:pipeline}
\end{figure}
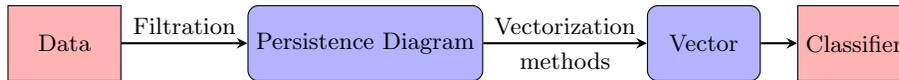

\begin{figure}[!ht]
    \centering

\begin{subfigure}[b]{0.45\textwidth}
    \centering
    \includegraphics[width=0.95\textwidth]{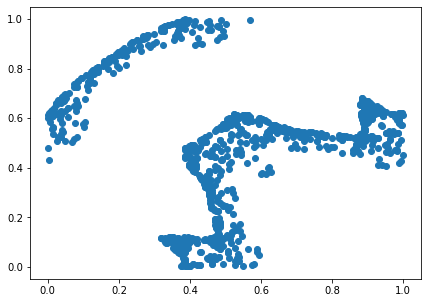}
     \caption{}
    \label{fig:Pip_applied:a}
    \end{subfigure}
    \begin{subfigure}[b]{0.45\textwidth}
     \centering
    \includegraphics[width=0.95\textwidth]{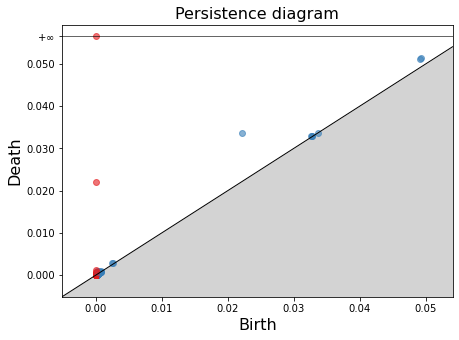}
     \caption{}
    \label{fig:Pip_applied:b}
    \end{subfigure}
    \begin{subfigure}[b]{0.45\textwidth}
    \centering
    \includegraphics[width=0.9\textwidth]{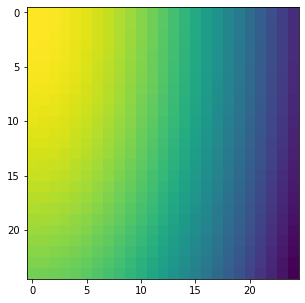}
    \caption{}
    \label{fig:Pip_applied:c}
    \end{subfigure}
    \begin{subfigure}[b]{0.45\textwidth}
     \centering
    \includegraphics[width=0.9\textwidth]{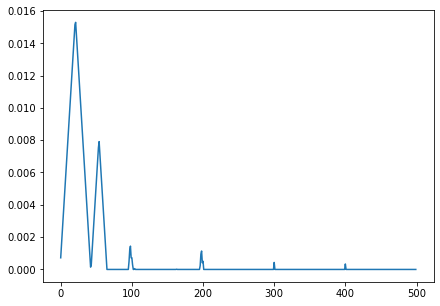}
     \caption{}
  \label{fig:Pip_applied:d}
   \end{subfigure}
    \begin{subfigure}[b]{0.45\textwidth}
     \centering
   \includegraphics[width=0.9\textwidth]{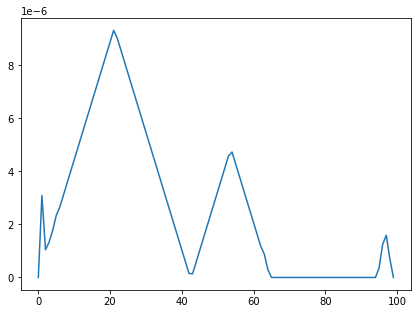}
     \caption{}
  \label{fig:Pip_applied:e}
   \end{subfigure}    
\begin{subfigure}[b]{0.45\textwidth}
     \centering
     \includegraphics[width=0.9\textwidth]{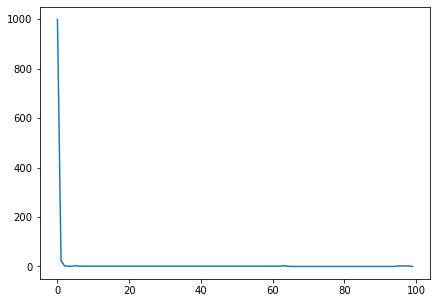}
    \caption{}
     \label{fig:Pip_applied:f}
    \end{subfigure}
    \caption{Pipeline application for a point cloud data \del{(left)} \rev{(Figure \ref{fig:Pip_applied:a})}. The persistence diagram associated \del{(right)} \rev{(Figure \ref{fig:Pip_applied:b})}. In the second \rev{ and third} row\rev{s}, four different vectorization methods for the same PD\rev{, namely Persistence Images (Figure \ref{fig:Pip_applied:c}), Persistence Landscapes (Figure \ref{fig:Pip_applied:d}), Persistence Silhouette (Figure \ref{fig:Pip_applied:e}) and Betti Curves (Figure \ref{fig:Pip_applied:f})}.}
    \label{fig:Pip_applied}
\end{figure}

The goal of this section is to define and describe the pipeline we use for the topological study of digital data in a machine learning context. As already mentioned, we rely heavily on the tools provided by persistent homology, which will be explored and discussed here. Figure \ref{fig:pipeline} shows the general scheme of our pipeline and its main elements. The digital data (Section \ref{3subsec1}) is filtered (Section \ref{3subsec2}) in order to generate a persistence diagram (Section \ref{3subsec3}). We vectorize the PD by means of a vectorization method (Section \ref{3subsec4}). Finally, the collection of all vectors is the input for a machine learning classifier (Section \ref{3subsec5}). Figure \ref{fig:Pip_applied} shows an application of the pipeline to \del{a} digital data.

\subsection{Data}\label{3subsec1}
In PH, data are often modelled as points in a metric space or as functions on some topological space. This is straightforward, as the most common digital data found in application\rev{s} are either $n$-dimensional vector\rev{s} or signals and images. Digital data are suitable for both modelling. For example, a $n \times n$ greyscale image can be viewed either as a function with domain a grid $\subseteq \mathbb{R}^2$ with values in $\mathbb{R}$, or as a vector (a point) in the space $\mathbb{R}^{n^2}$. We want to emphasize, however, that the way in which we model data is of fundamental importance for the study that follows. This is mainly due to the type of metrics that the data inherit from the two modelling. Continuing our example, the classical metric between two points of $\mathbb{R}^{n^2}$ is the Euclidean one, while between two functions we have $\| \cdot \|_2$, $\| \cdot \|_\infty$ or others. Although the metrics can easily be changed according to the context, the different modelling has profound repercussions both mathematically and in terms of application. There is a very rich literature on both the modelling data as points in a metric space \cite{carlsson2009topology,Epstein2011TopologicalDA,10.1145/1057432.1057449} and modelling as functions \cite{frosini_1990,biasotti2008multidimensional,bergomi2019towards}. Depending on the type of data, one modelling may be preferred to another. Let us now describe the digital data types we studied in Section \ref{sec4} and the associated preferred modelling.

\subsubsection{Point cloud data}\label{3subsubsec1}
Point clouds data are finite metric spaces $(X,\delta)$. We recall that finite metric spaces are naturally equipped with the discrete topology, and the topological dimension of any discrete space is $0$. Since every finite metric space is discrete, point clouds do not inherit a (not trivial) topology. As a consequence, the homology of a point cloud is always trivial, with the exception of $H_0$. As finite metric spaces, the natural modelling of this type of data is that of a vector embedded in a larger Euclidean space $(Y, d)$. Typically $Y = \mathbb{R}^n$ and $d$ is the Euclidean metric, but this model is suitable for generalization. In Section \ref{sec4} \rev{we} will always \del{be} \rev{have} $Y = \mathbb{R}^n$.

\subsubsection{Images}\label{3subsubsec2}
A digital image is an image composed of pixels. In standard 8-bit greyscale images, each pixel has a discrete value ranging from $0$ to $255$ representing its greyscale intensity. Although it can be considered as a vector of dimension $n\cdot m$, where the greyscale image has size $n \times m$, modelling an image as a function is preferable. This kind of modelling takes into account the fact that only close pixel\rev{s} can be connected, not also distant ones. Therefore a digital image is a function \del{between} \rev{from} a grid $G \subseteq \mathbb{R}^2$ to a suitable finite interval of $\mathbb{R}^n$, depending on the number of channels of the image. The most common number of channel\rev{s} is $1$, for greyscale, or $3$, for color images. However, there \del{are} \rev{exist} additional image channel encoding\rev{s} available.

\subsubsection{Graphs}\label{3subsubsec3}
Graphs are structures made by a set of vertices which are connected by edges. There is a distinction between undirected graphs, where the edges link two vertices symmetrically, and directed graphs, where the link is not symmetrical. Another distinction is between weighted and unweighted graphs. More specifically, a graph is an ordered pair $G=(V, E)$ where $V$ is the set of vertices and $E\subseteq\left\{(x, y) | x, y \in V\right\}$. Typically $(x, y) \in E \Rightarrow x \neq y$ but this is not always the case. A graph is seen as a function with domain $E$ and codomain $\mathbb{R}$.

\subsection{Filtrations}\label{3subsec2}
Persistent homology examines the shape of the data at different scales. As already mentioned in Section \ref{2subsec2}, this means that at each scale $j$ the data is represented as a simplicial complex $\mathcal{K}_j$. Formally, a \emph{filtration} of a simplicial complex $\mathcal{K}$ is a finite sequence of nested subcomplexes
\[ \emptyset \subseteq \mathcal{K}_0 \subseteq \mathcal{K}_1 \subseteq \dots \subseteq \mathcal{K}_n = \mathcal{K}. \]
The inclusion $\mathcal{K}_i \hookrightarrow \mathcal{K}_j$ induces a homomorphism $f_p^{i,j}\colon H_p(\mathcal{K}_i)\to H_p(\mathcal{K}_j)$ of the simplicial homology groups for each dimension $p$ and each $0\leq i\leq j \leq n$. Usually these homomorphisms are actually isomorphisms. In these cases\rev{,} no topological events have occured between time $i$ and time $j$ and, more importantly, Betti numbers do not change. However, there are cases in which these homomorphisms are not injective or surjective. In these cases, topological events occur, which is precisely what we are interested in. We say that a new topological feature is born at time $j$ when the homomorphism $f_p^{i,j}$ is not surjective. We say that a topological feature dies at time $j$ when the homomorphism $f_p^{i,j}$ is not injective. We keep track of the pairs (birth, death) of the topological features and collect them in the so\rev{-}called \rev{P}ersistence \rev{D}iagram (PD). \newline

We would like to focus on the importance of filtrations in our study, as the choice of the filtration is the first fundamental step in a topological study of the data. There are several filtrations available, and they are mainly related to the choice of data modelling. We want to stress that different filtrations yield different topological features, and thus possibly very different PDs. Moreover, some filtrations enable to emphasize points with greater or lesser persistence, or prevent the creation of certain features where there should be none. Therefore, the choice of filtration should be made with caution.

\subsubsection{Filtration for point clouds}\label{3subsubsec4}
The alpha complex is a way of forming abstract simplicial complexes from a set of points. Hence, it represents an ideal filtration for point clouds. Let $(X,\delta)$ be a point cloud embedded in a larger metric space $(Y,d)$. The elements of $X$ are the vertices of the alpha complex. Fixed a real parameter $\alpha > 0$, we define $$A_x^\alpha := B(x, \alpha) \cap \left\{y \in Y : d(y,x) \leq d(y, \tilde{x}) \text{ for every } x\neq \tilde{x} \in X\right\}.$$ That is, we grow balls with radius $\alpha$ centered in each point of $X$, intersected with the Voronoi cell of each point. When $n$ \rev{sets} $A_{x_i}^\alpha$ intersect, a $(n-1)$-simplex is added to the simplicial complex. The growing of $\alpha$ naturally induces a filtration. For more information on alpha complex, see \cite{akkiraju1995alpha}. We point out that, by the nerve lemma, the alpha complex is homotopy equivalent to the union of the balls and also to the Čech complex. The advantage over the Čech complex is that it is significantly smaller, thus reducing the computational cost.

\subsubsection{Filtration for images}\label{3subsubsec5}
\begin{figure}[!ht]
\centering
    \begin{tikzpicture}[line cap=round,line join=round,>=triangle 45,x=1cm,y=1cm]
        \draw [line width=2pt] (0,6) -- (4,6);
        \draw [line width=2pt] (4,6) -- (4,2);
        \draw [line width=2pt] (4,2) -- (0,2);
        \draw [line width=2pt] (0,2) -- (0,6);
        \draw [line width=2pt] (0,4) -- (4,4);
        \draw [line width=2pt] (2,6) -- (2,2);
        \draw [line width=2pt] (4,2) -- (0,6);
        \draw [line width=2pt] (2,6) -- (4,4);
        \draw [line width=2pt] (2,2) -- (0,4);
        \begin{scriptsize}
            \draw [fill=qqqqff] (4,2) circle (2.5pt);
            \draw [fill=qqqqff] (2,2) circle (2.5pt);
            \draw [fill=qqqqff] (0,2) circle (2.5pt);
            \draw [fill=qqqqff] (0,4) circle (2.5pt);
            \draw [fill=qqqqff] (2,4) circle (2.5pt);
            \draw [fill=qqqqff] (4,4) circle (2.5pt);
            \draw [fill=qqqqff] (4,6) circle (2.5pt);
            \draw [fill=qqqqff] (2,6) circle (2.5pt);
            \draw [fill=qqqqff] (0,6) circle (2.5pt);
        \end{scriptsize}
    \end{tikzpicture}
\caption{Pixel connection in cubical complexes.}
\label{fig:Cubical}
\end{figure}
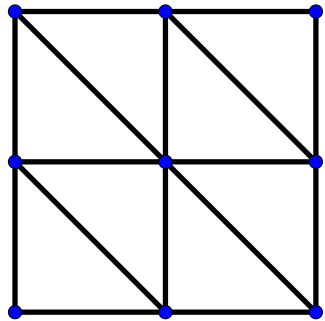

The alpha complex is still a viable filtration for images, since they can be interpreted as vectors. However, other approaches that make greater use of the fixed structure of the image are preferable. In particular, the cubical complex is the ideal filtration as it exploits two key features of images. The first feature is that not all pixels should be connected with each other, but only with the neighbours. Figure \ref{fig:Cubical} shows the possible connection between pixels in a cubical complex. The second feature is related to the modelling of images as functions. In contrast to the alpha complex, where all points were immediately inserted as vertices of the simplicial complex, a pixel becomes a vertex of the simplicial complex only when its intensity becomes greater than a certain threshold value $t$. Similarly, a $1$-symplex is added only if two adjacent pixels (in the sense of Figure \ref{fig:Cubical}) both have intensities greater than $t$. The same applies to the $2$-simplexes. As $t$ increases, we obtain a filtration. More formally, an elementary cube is any translate of a unit cube $[0,1]^n$ embedded in Euclidean space $\mathbb{R}^m$, for some $n, m \in \mathbb{N}$ with $n \leq m$. A set $I \subset \mathbb{R}^m$ is a cubical complex if it is homeomorphic to a\del{n} union of elementary cubes. For more information about cubical complexes, we refer the reader to \cite{kaczynski2004computational}.

\subsubsection{Filtration for graphs}\label{3subsubsec10}
Graphs are particular structures which require\del{s} \emph{ad hoc} filtrations. In particular, graphs share many similarities with point cloud\rev{s}, but with few key differences. The main difference is that vertices may not be connected to each other. Therefore, no matter how much you increase the filtration value, some $1$-simplexes will never be created. This of course impacts also higher-dimensional simplexes. Another difference is that all vertices form $0$-simplexes, but their filtration value may not be $0$. For example,  the number of incident edges of every vertex could be used as filtration value. There are many possible filtrations associated with graphs, depending mainly on the type of graph considered. In any case, the difference lies essentially in the filtration value associated with each simplex, not in the creation of the simplexes themselves. For this reason, we now describe how the simplexes enter the filtration, postponing the description of the filtration value chosen to the section \ref{4subsec4}. Every vertex form a $0$-simplex. If two vertices are connected by an edge, a $1$-simplex is formed. Similarly, when three vertices are pairwise linked, a $2$-simplex is formed. In general, a clique is a subset of $V$ whose vertices are all pairwise connected. Each clique of $k$ vertices form a $(k-1)$-simplex.

\subsection{Persistence diagrams}\label{3subsec3}
Persistence diagrams are the collection of pairs (birth, death) of topological features emerged by filtering a simplicial complex. We refer to this pairs as $q_j = (b_j, d_j)$, where $b_j$ is the birth of the $j$th $k$-dimensional hole and $d_j$ its death. Mathematically, this collection is a multiset, that is a set in which same elements can appear multiple times. For further details on persistence diagrams, we refer the reader to \cite{biasotti2008describing}, \cite{carlsson2009theory}, and \cite{edelsbrunner2008persistent}. Each pairs has a multiplicity $\mu(q_j)$ indicating how many holes share both the birth and the death time. The points $(t, t)$ of the diagonal are added to the dipersistence diagram with infinite multiplicity for technical reasons. Since the death of a topological feature occurs at a larger time its birth, PDs are multiset over the set $\bar{\Delta}^* := \left\{(x,y) \in \mathbb{R}^2 : x \leq y\right\} \cup \left\{(x,\infty) : x \in \mathbb{R}\right\}$. It holds that $\mu(q) = 0$ if and only if $q \not\in D$, where $q \in \bar{\Delta}^*$ and $D$ is a persistence diagram. The equality $\ell(q) = k$ means that the point $q \in D$ corresponds to a feature in $H_k$.  We can equip the space of persistence diagrams with the \textit{bottleneck distance} (also called \textit{matching distance})
\[ W_\infty(D, D') := \inf_{\varphi \colon D \to D'} \sup_{q \in D} \| q - \varphi(q)\|_\infty, \]
where $D, D'$ are persistence diagrams and $\varphi$ is a bijection from $D$ to $D'$. Another popular metric in the space of PDs is the $p$-Wasserstein distance, defined as
\[W_p(D, D') := \inf_{\varphi \colon D \to D'}\left[\sum_{q \in D}\left(\| q - \varphi(q)\|_\infty\right)^p\right]^{\frac{1}{p}}, \quad p\geq 1. \]
For more information on bottleneck and Wasserstein distance, we refer the reader to \cite{cohen2007stability} and \cite{edelsbrunner2008persistent}. We point out that in the general case there exists a bijection $\varphi \colon D \to D'$ only because we have added the points in the diagonal with infinite multiplicity. The most important property of PDs is their stability. That is, a small perturbation of the simplicial complex yields a small perturbation of the associated PD. This property is of fundamental importance in applications as it guarantees robustness against noise and repeatability. Multisets lack fundamental mathematical and statistical properties required in a machine learning context and therefore they cannot be directly \del{applied in} \rev{processed by} an ML algorithm. To give an example, the mean of two multiset\rev{s} is not well defined. Suitable transformation of PDs into objects that enjoy excellent mathematical properties and can be used in ML is needed. These transformations are called \textit{representation methods} or \textit{featurization methods}. We limit our study to the \textit{vectorization methods}, \rev{i.e. procedures for transforming a persistence diagram into a vector.} \del{ that is a  transformation of a persistence diagram in a vector.} Additional representation methods are currently available, i.e. kernel methods. Due to the high computational cost of these methods, they have been omitted from this work, but the interested reader can find more information in \cite{chen2019topological}, \cite{pun2018persistent} and \cite{corbet2019kernel}.

\subsection{Vectorization methods}\label{3subsec4}
A vector representation of a PD consists of an embedding of the space of PD in a vector space or, more generally, in a Hilbert space. The fundamental requirement of this embedding is stability, i.e., that small perturbations of the PD correspond to small perturbations of the associated vector. Various embeddings are obviously possible, and each of them will define a different vector representation of the PD. The vector representations presented in this work are all stable with respect to the bottleneck distance or $1$-Wasserstein distance, with the exception of Betti curves (see Section \ref{3subsubsec9}). Finally, we want to stress the fact that all these vectors live in a vector space and thus enjoy mathematical and statistical properties that were not available to the space of multisets. Hence, they can be directly introduced in a machine learning method. Nonetheless, different representations of the same PD yield different vectors, with possible very different results in a\rev{n} ML algorithm. We point out that this is precisely the main goal of this work, \rev{to} find a correlation between task, filtration and representation. All the subsequent methods require\del{s} a change of coordinates of the PD. Henceforth, unless otherwise specified, a point $q \in D$ will have coordinates $q = (\frac{b+d}{2}, \frac{d-b}{2})$, where $(b, d)$ are the usual birth and death of a topological feature. We point out that in $H_0$ there is always a connected component that never dies. Since these methods do not handle \rev{the} infinite persistence of some points, we replace the infinite value by a very large one in relation to the other persistences obtained.
Each of these methods is derived from the Gudhi library. For more information about Gudhi and the Python implementation of these methods, we refer the reader to \cite{gudhi:urm}.

\subsubsection{Persistence image}\label{3subsubsec6}
\begin{figure}[!ht]
     \centering
     \begin{subfigure}[b]{0.24\textwidth}
        \centering
         \includegraphics[width=0.95\textwidth]{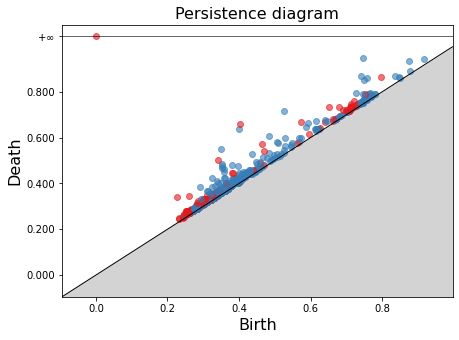}
         \caption{}
         \label{fig:PI:a}
     \end{subfigure}
      \begin{subfigure}[b]{0.24\textwidth}
        \centering
         \includegraphics[width=0.95\textwidth]{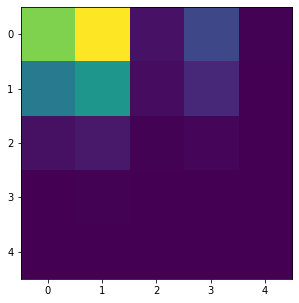}
         \caption{}
         \label{fig:PI:b}
     \end{subfigure}
          \begin{subfigure}[b]{0.24\textwidth}
        \centering
         \includegraphics[width=0.95\textwidth]{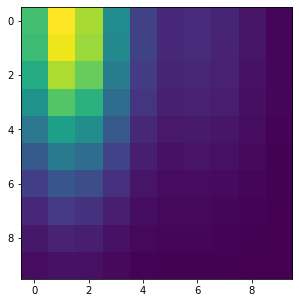}
         \caption{}
         \label{fig:PI:c}
     \end{subfigure}
          \begin{subfigure}[b]{0.24\textwidth}
        \centering
         \includegraphics[width=0.95\textwidth]{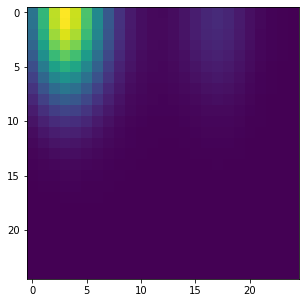}
         \caption{}
         \label{fig:PI:d}
     \end{subfigure}
     \caption{Persistence diagram (\rev{Figure \ref{fig:PI:a}}) and three persistence images for $(\sigma, n) = (0.1, 5), (0.1, 10), (0.05, 25)$ respectively \rev{ in Figures \ref{fig:PI:b}, \ref{fig:PI:c}, \ref{fig:PI:d}}.}
     \label{fig:PI}
\end{figure}

\rev{A} Persistence \rev{I}mage (PI) is a finite-dimensional vector representation of persistence diagrams. For more information on PI, we refer the reader to \cite{adams2017persistence}. This method basically divides the PD domain into an $n\times n$ grid and, for each point $q$ of the PD, defines a Gaussian centered in $q$ with variance $\sigma$. It returns an $n\times n$ image where the intensity of each pixel is given by the sum of the values of all Gaussians at that point in the grid, weighted by an appropriate function $f$ that must be $0$ on the diagonal, continuous and piecewise differentiable. Denoting with $m$ the persistence value of the most persistent feature, the weight function is
\[ f(t) := \begin{cases}
0 \text{ if } t\leq 0,\\
\frac{t}{m} \text{ if } 0 < t < m,\\
1 \text{ if } t\geq m.
\end{cases} \]
The parameters of the method are $n$ and $\sigma$ and will be selected by grid search. Figure \ref{fig:PI} shows a persistence diagram\del{s} and three different PIs for various parameters. It is evident how different parameter values greatly influence the resulting image. In our pipeline we determine the optimal parameters with a grid search approach between the following setup: $\sigma \in \left\{0.1, 1, 10\right\}, n \in \left\{5, 10, 25\right\}$.

\subsubsection{Persistence landscape}\label{3subsubsec7}
\begin{figure}[!ht]
     \centering
     \begin{subfigure}[b]{0.24\textwidth}
        \centering
         \includegraphics[width=0.95\textwidth]{PD.png}
         \caption{}
         \label{fig:PL:a}
     \end{subfigure}
      \begin{subfigure}[b]{0.24\textwidth}
        \centering
         \includegraphics[width=0.95\textwidth]{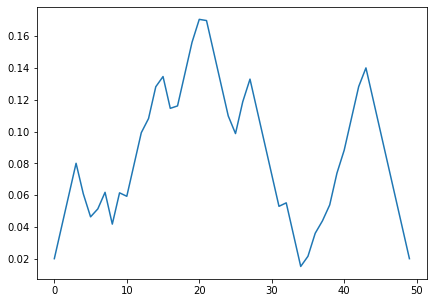}
         \caption{}
         \label{fig:PL:b}
     \end{subfigure}
          \begin{subfigure}[b]{0.24\textwidth}
        \centering
         \includegraphics[width=0.95\textwidth]{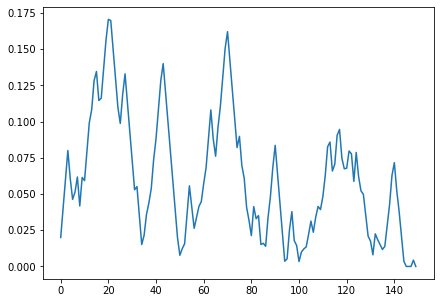}
         \caption{}
         \label{fig:PL:c}
     \end{subfigure}
          \begin{subfigure}[b]{0.24\textwidth}
        \centering
         \includegraphics[width=0.95\textwidth]{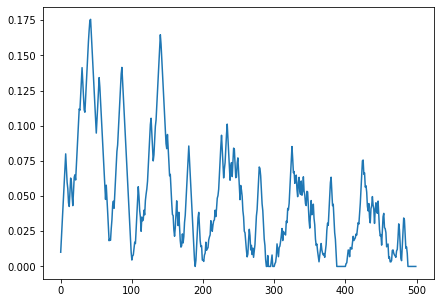}
         \caption{}
         \label{fig:PL:d}
     \end{subfigure}
     \caption{Persistence diagram (\rev{Figure \ref{fig:PL:a}}) and three persistence landscapes for $(n, r) = (1, 25), (3, 50), (5, 100)$ respectively \rev{ in Figures \ref{fig:PL:b}, \ref{fig:PL:c}, \ref{fig:PL:d}}.}
     \label{fig:PL}
\end{figure}

\rev{The} Persistence \rev{L}andscape (PL) is another method of vector representation of PD that enjoys excellent statistical properties introduced in \cite{bubenik2015statistical}. In particular, the PL is a function that live\rev{s} in a vector space, a great mathematical environment for working with ML. More formally, PLs are piecewise constant functions $\lambda \colon \mathbb{N} \times \mathbb{R} \to \overline{\mathbb{R}}$. To define $\lambda$, we tent each persistence point $q = (\frac{b+d}{2}, \frac{d-b}{2}) \in D$ to the baseline $x = 0$ with the following function
\[ \Lambda_q(t) := \begin{cases}
t - b \text{ if } t \in [b, \frac{b+d}{2}], \\
d - t \text{ if }t \in (\frac{b+d}{2}, d], \\
0 \text{ otherwise.}
\end{cases} \]
The persistence landscapes of $D$ is the collection of such functions
\[ \lambda_D(k, t) := \kmax\limits_{q \in D}\Lambda_q(t), \quad k \in \mathbb{N}, t \in [0, T], \]
where $\kmax$ is the $k$-th largest value in the set and $T$ is a real number such that $d \leq T$ for any death time $d$ of a topological feature. Since $\lambda$ is piecewise constant, it can be discretized by looking only at the point \del{in} \rev{at} which it change\rev{s} value. This discrete function is the vector representation of the PD. A Central limit theorem for PLs holds. The parameters for persistence landscapes are the number of landscapes considered $n$ and the discretization resolution $r$. In our grid search approach, we consider the following setup: $n=5, r \in \left\{25, 50, 75, 100\right\}$. Figure \ref{fig:PL} shows a PD and three persistence landscapes associated.

\subsubsection{Persistence silhouette}\label{3subsubsec8}
\begin{figure}[!ht]
     \centering
     \begin{subfigure}[b]{0.24\textwidth}
        \centering
         \includegraphics[width=0.95\textwidth]{PD.png}
         \caption{}
         \label{fig:PS:a}
     \end{subfigure}
      \begin{subfigure}[b]{0.24\textwidth}
        \centering
         \includegraphics[width=0.95\textwidth]{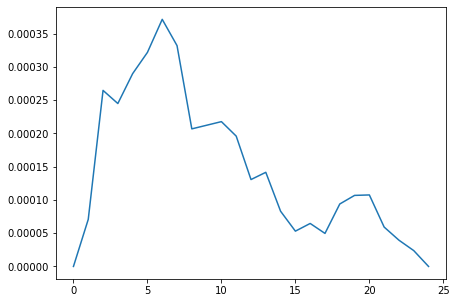}
         \caption{}
         \label{fig:PS:b}
     \end{subfigure}
          \begin{subfigure}[b]{0.24\textwidth}
        \centering
         \includegraphics[width=0.95\textwidth]{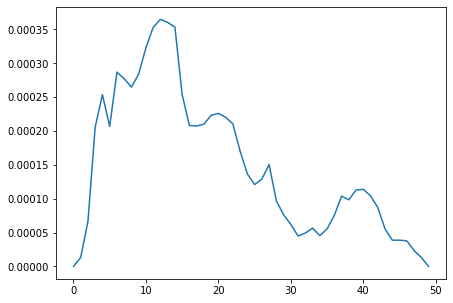}
         \caption{}
         \label{fig:PS:c}
     \end{subfigure}
          \begin{subfigure}[b]{0.24\textwidth}
        \centering
         \includegraphics[width=0.95\textwidth]{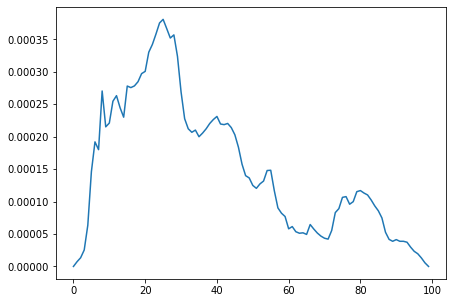}
         \caption{}
         \label{fig:PS:d}
     \end{subfigure}
     \caption{Persistence diagram (\rev{Figure \ref{fig:PS:a}}) and three persistence silhouette for $r = 25, 50, 100$ respectively \rev{ in Figures \ref{fig:PS:b}, \ref{fig:PS:c}, \ref{fig:PS:d}}.}
     \label{fig:PS}
\end{figure}

\rev{The} Persistence \rev{S}ilhouette (PS) is a method of vector representation of PD with the same core idea of PL. More specifically, it is a piecewise constant function defined as
\[ \phi(t) := \frac{\sum_{j=1}^m w_j\Lambda_j(t)}{\sum_{j=1}^m w_j}, \]
where $m$ is the number of off\rev{-}diagonal points, $w_j$ is a weight and $\Lambda_j$ is the same functions as in Section \ref{3subsubsec7}. Again, the vector representation of the PD comes from the discretization of $\phi$. In our pipeline, we use constant weight $w_j =1$ for every $j=1, \dots,m$. The only parameter of the method is the resolution $r$ and the grid search takes values $r\in\left\{25, 50, 75, 100\right\}$. Figure \ref{fig:PS} shows a PD and three persistence silhouettes. We point out the similarity of the vectors with different parameters values. For more information on Persistence Silhouette, we refer the reader to \cite{chazal2014stochastic}.

\subsubsection{Betti curve}\label{3subsubsec9}
\begin{figure}[!ht]
     \centering
        \begin{subfigure}[b]{0.24\textwidth}
        \centering
         \includegraphics[width=0.95\textwidth]{PD.png}
         \caption{}
         \label{fig:BC:a}
     \end{subfigure}
      \begin{subfigure}[b]{0.24\textwidth}
        \centering
         \includegraphics[width=0.95\textwidth]{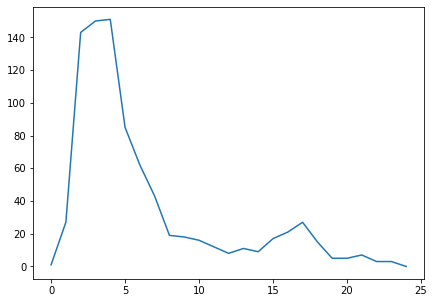}
         \caption{}
         \label{fig:BC:b}
     \end{subfigure}
          \begin{subfigure}[b]{0.24\textwidth}
        \centering
         \includegraphics[width=0.95\textwidth]{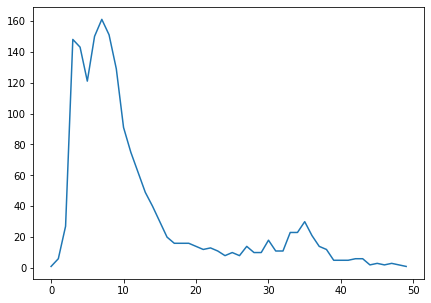}
         \caption{}
         \label{fig:BC:c}
     \end{subfigure}
          \begin{subfigure}[b]{0.24\textwidth}
        \centering
         \includegraphics[width=0.95\textwidth]{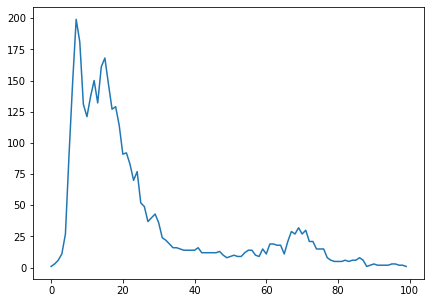}
         \caption{}
         \label{fig:BC:d}
     \end{subfigure}
     \caption{Persistence diagram  (\rev{Figure \ref{fig:BC:a}}) and three \rev{B}etti curves for $r = 25, 50, 100$ respectively \rev{ in Figures \ref{fig:BC:b}, \ref{fig:BC:c}, \ref{fig:BC:d}}.}
     \label{fig:BC}
\end{figure}

\rev{The} Betti curve (BC) is yet another vectorization method for persistence diagrams presented in \cite{umeda2017time}. The \rev{B}etti curve are a $\mathbb{Z}$-indexed family of functions defined as $\beta_z\colon \mathbb{R} \to \mathbb{R}, \beta_z(t) := \#\left\{q = (b, d) \in D : \ell(q) = z \text{ and } b \leq t \leq d\right\}$, where $\ell(q) = z$ means that $q$ is a topological event in the homology group $H_z$. The function is then vectorized over a uniform grid of a closed interval and resolution $r$. For those familiar with persistence barcodes, this method informally counts the number of bars present in the persistence barcodes at any given time, after an appropriate  persistence normalization. The resolution $r$ in our pipeline takes values in the grid $r \in \left\{25, 50, 75, 100\right\}$. Figure \ref{fig:BC} shows a PD and three BCs for different resolution\rev{s} of the grid. \newline

In a persistence diagram\rev{,} all points of the different homology dimensions are present. However, some of these dimensions may be more or less suitable for the given study. For this reason, our pipeline follows three approaches. The first approach is to consider each homology group individually. We highlight the fact that it may happen that $H_i$ of some data are empty, for $i \neq 0$, while others are non-empty. In such cases, we replace the empty $H_i$ \del{by} \rev{with} the single point $(0,0)$. In the following sections, this approach will be referred as $H_i$ approach, for $i$ varying in the homology groups available. The second approach is to forget about the homology dimension of the points, referred \rev{to} as the `fused' approach, whereas the third is to concatenate the vector representation of the PD for each homology dimension, referred as the `concat' approach. \newline

We want to emphasize here that there is no homogeneity \del{on} \rev{in} the number of parameters used in the grid search of the various vectorization methods. In particular, persistence images have a total of nine parameter combinations, while the others only have four. This is done on the one hand to maintain a comparison with other works, on the other hand\rev{,} because we are not interested in finding the best vector representation, but rather in finding links between data and vectorization methods. Moreover, some methods are more flexible than others, and therefore more parameter\rev{-}dependent. Hence\rev{,} in these cases\rev{,} it is more suitable to test a larger number of parameters. Finally, we stress that the purpose of our study is to verify the usefulness of a topological study of the data and to investigate the preferred vector representation in certain contexts. The stability of the representation is of fundamental importance, as it guarantees a stable synergy between TDA and ML.

\subsection{Machine Learning classifiers}\label{3subsec5}
At this point\rev{,} we are finally able to introduce the vectorization of the PDs in \del{a} machine learning classifiers. We want to emphasize that the pipeline does not perform \del{a} parameter tuning either for representation methods or for classifiers. We would also point out that the vector generated from the previous section may have very different sizes and necessarily different computational costs, but this aspect is not taken into account in our experiments. In the pipeline, a total of nine classifiers are trained for each representation. Such classifiers are: \rev{S}upport \rev{V}ector \rev{C}lassifiers \rev{SVC} \del{(}with kernel RBF and C=$\{1, 2, 3, 5, 10, 20\}$\del{)}; a random forest classifier ($\#$trees = $100$); and Lasso ($\alpha = 1$).

These classifiers are very standard in machine learning literature and we have reported the parameters used for each of them. For more details, we refer the reader to \cite{cortes1995support,breiman2001random,hoerl1970ridge}, and  \cite{tibshirani1996regression}. Each of these classifiers is derived from the Scikit-learn library. For more information about Scikit-learn and the Python implementation of these methods, we refer the reader to \cite{scikit-learn}. We want to emphasize that there are such a large number of SVCs just for comparison with other works. Each of the classifiers will perform a $10$-fold cross validation \cite{doi:10.1080/00401706.1974.10489157} on the data and we will report the best result obtained for each run, along with the representation method that achieved it. We stress the fact that for each run we report the best accuracy result achieved, regardless of the vectorization method or classifier that achieved it. For completeness, we also provide a table with the (mean) accuracy result of the best single method at the end of each experiment. \rev{Finally, a statistical analysis of the results obtained during the experiments was carried out, and presented in Section \ref{sec5}.} 

\subsection{Further improvements}\label{3subsec6}
The pipeline described up to this point is the basic version of our procedure. 
\rev{It is worth pointing out that this pipeline may have high requirements in terms of time and memory: for example, our pipeline needs a few seconds to classify a sample dynamic trajectory, while could take about an hour for classifying a complex sample of a collaboration network.}
This is a basic version because we are omitting some possible improvements in the two main contexts of the pipeline, TDA and ML. For example, multiple filtrations could be available for the same type of data, or more advanced ML methods such as \textit{ensemble learning} could be used. Since our goal is to take a first step in the study of the best PD representations for different types of digital data, these improvements are initially omitted from the pipeline. If, however, the results obtained are not satisfactory in terms of accuracy and/or stability, suitable improvements will be made. In particular, we stress the fact that the stability of the representation is our most important concern and high accuracy results without stability are not satisfactory. Necessarily, however, poor but stable results will be equally unsatisfactory. These adjustments to the basic pipeline will be discussed in detail when introduced, as they are strongly linked to the type of dataset considered.

\section{Results}\label{sec4}
In this section\rev{,} we are going to apply the pipeline presented in Section \ref{sec3} to different heterogeneous types of datasets. The goal of this section is to discuss the usefulness of a topological study of the data. This is accomplished by showcasing the excellent accuracy results achieved by the pipeline. All results reported here refer to test accuracy. For each dataset, we perform a $10$-split cross\rev{-}validation with an $80\%$ training data and $20\%$ test data, and report the results of the pipeline over the course of the ten runs, along with mean accuracy and standard deviation. Finally, since when we have a never-seen data we would like to know how to perform classification, we also report the accuracy value of the single best method for each dataset (i.e. the best combination \{vectorization, classifier\}). 

\subsection{Dynamic dataset}\label{4subsec1}
\begin{figure}[!ht]
     \centering
     \begin{subfigure}[b]{0.19\textwidth}
        \centering
         \includegraphics[width=0.95\textwidth]{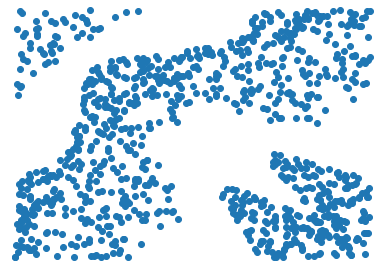}
         \caption{}
         \label{fig:Dyn_par:a}
     \end{subfigure}
     \begin{subfigure}[b]{0.19\textwidth}
        \centering
         \includegraphics[width=0.95\textwidth]{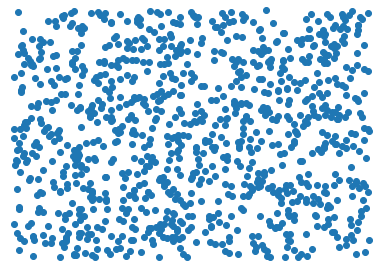}
         \caption{}
         \label{fig:Dyn_par:b}
     \end{subfigure}
     \begin{subfigure}[b]{0.19\textwidth}
        \centering
         \includegraphics[width=0.95\textwidth]{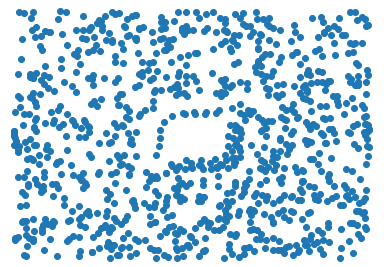}
         \caption{}
         \label{fig:Dyn_par:c}
     \end{subfigure}
     \begin{subfigure}[b]{0.19\textwidth}
        \centering
         \includegraphics[width=0.95\textwidth]{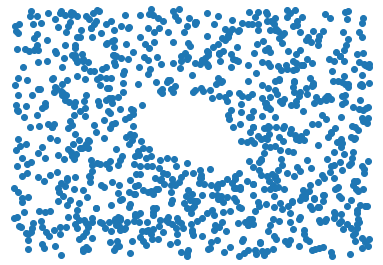}
         \caption{}
         \label{fig:Dyn_par:d}
     \end{subfigure}
     \begin{subfigure}[b]{0.19\textwidth}
        \centering
         \includegraphics[width=0.95\textwidth]{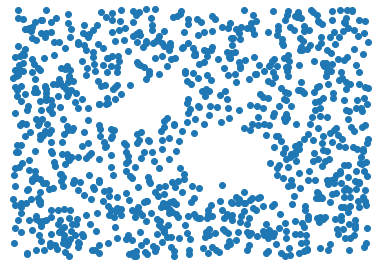}
         \caption{}
         \label{fig:Dyn_par:e}
     \end{subfigure}
     \caption{Example of truncated orbits $\{(x_n,y_n), n=0,\dots, 1000\}$ for the first $1000$ iterations of the linked twisted map for $r=[2, 3.5, 4, 4.1, 4.3]$ respectively.}
     \label{fig:Dyn_par}
\end{figure}

\begin{figure}[!ht]
     \centering
      \begin{subfigure}[b]{0.19\textwidth}
        \centering
         \includegraphics[width=0.95\textwidth]{Dynamic_43.png}
         \caption{}
         \label{fig:Dyn_start:a}
     \end{subfigure}
     \begin{subfigure}[b]{0.19\textwidth}
        \centering
         \includegraphics[width=0.95\textwidth]{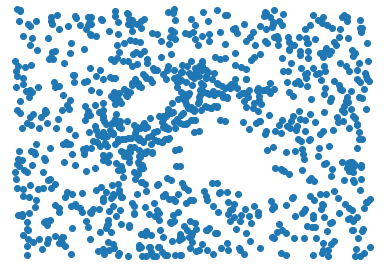}
         \caption{}
         \label{fig:Dyn_start:b}
     \end{subfigure}
     \begin{subfigure}[b]{0.19\textwidth}
        \centering
         \includegraphics[width=0.95\textwidth]{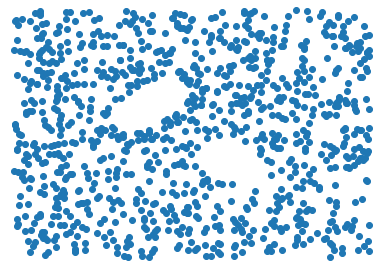}
         \caption{}
         \label{fig:Dyn_start:c}
     \end{subfigure}
     \begin{subfigure}[b]{0.19\textwidth}
        \centering
         \includegraphics[width=0.95\textwidth]{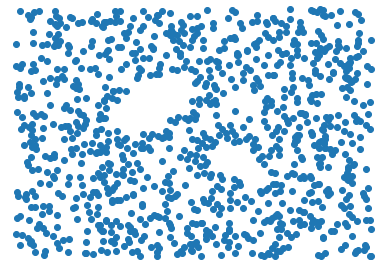}
         \caption{}
         \label{fig:Dyn_start:d}
     \end{subfigure}
     \begin{subfigure}[b]{0.19\textwidth}
        \centering
         \includegraphics[width=0.95\textwidth]{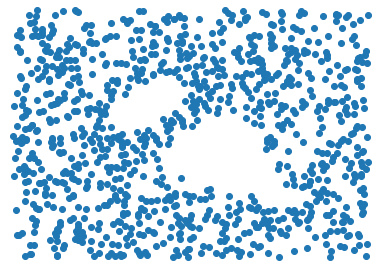}
         \caption{}
         \label{fig:Dyn_start:e}
     \end{subfigure}
     \caption{Example of truncated orbits $\{(x_n,y_n), n=0,\dots, 1000\}$ for the first $1000$ iterations of the linked twisted map for $r=4.3$ and five different starting points.}
     \label{fig:Dyn_start}
\end{figure}

The first dataset we describe comes from data arising from a discrete dynamical system modeling fluid flow. The dynamical system here presented is a linked twisted map, that is a Poincaré section of an eggbeater-type flow. A Poincaré section is the discretization of a continuous dynamical system obtained by following the path of a particle's location at discrete time intervals. The equations for the linked twisted map are\rev{:}
\begin{align*}
    \begin{cases}
        x_{n+1} = x_n + ry_n(1-y_n) & \text{mod} 1\\
        y_{n+1} = y_n + rx_{n+1}(1-x_{n+1}) & \text{mod} 1
    \end{cases}
\end{align*}
where $r$ is a positive parameter. For different values of the parameter $r$, different orbits $\{(x_n,y_n), n\in\mathbb{N}\}$ are generated. In some cases\rev{,} such orbits are dense in the domain $[0,1]^2$, in other cases\rev{,} voids occur. The task of this application is to classify the value of the parameter $r$ based on the orbit. In Figure \ref{fig:Dyn_par} five different orbits generated by the same starting point for five different values of the parameter $r$ are shown. It is clear how the value of $r$ strongly influences the orbit, and in particular the formation of voids. In Figure \ref{fig:Dyn_start} five orbits for the same parameter $r$ with different starting points are shown, clarifying that the shape of the orbit depends only on the value of the parameter $r$ and not on the starting point. This dataset is inspired by \cite{adams2017persistence} and \cite{chung2021persistence} and is composed of five different values of the parameter $r=[2, 3.5, 4, 4.1, 4.3]$, each with $50$ orbits generated from different random starting points, for a total of $250$ orbits. Each orbit consists of $1000$ points generated from $1000$ iteration of the linked twisted map. We split the dataset in $80\%$ train set and $20\%$ test set. The dataset is naturally a point cloud, therefore the alpha filtration is best suited to generate the PDs. We recall that the pipeline performs a grid search between the most classical representation techniques for persistence diagrams, such as persistence images or persistence landscapes, for different values of the parameters. Such methods are evaluated with a ten\rev{-}split cross\rev{-}validation and the best results are returned. We report in Table \ref{tab:Dynamic_runs} the accuracy results obtained by the pipeline for each of the ten runs, alongside the abbreviation of the best representation method (PI for persistence image and so on). It is worth noticing that the results show a good consistency of the best representation method. As already stated, consistency is of great importance to us as it demonstrates the feasibility of a single method for classification. The last row report\rev{s} the mean $\pm$ standard deviation of the pipeline over the course of the ten runs. Finally, Table \ref{tab:Dynamic_best} report\rev{s} the best mean accuracy of the best single method, that is the combination vectorization method and classifier. Of course, the accuracy of a single method is lower than the mean in Table \ref{tab:Dynamic_runs}, since in that case the best accuracy could be achieved by different methods in the various runs. In conclusion, the results obtained by the pipeline in the dynamic dataset are very satisfactory, both from the point of view of accuracy and consistency of the representations.

\begin{table}[!ht]
{\footnotesize
    \caption{Accuracy for \rev{the} dy\rev{n}amical \rev{system} dataset.}\label{tab:Dynamic_runs}
    \begin{center}
    \begin{tabular}{ccccc} \hline
        Accuracy:   &   $H_0$           &   $H_1$           &   $H_0+H_1$ (fused)   &   $H_0+H_1$ (concat) \\
        \hline
        Run $1$     &   $0.493$(PI)     &   $0.960$(PL)     &   $0.507$(PI)         &   $0.920$(PL) \\
        Run $2$     &   $0.480$(PI)     &   $0.880$(PL)     &   $0.600$(PI)         &   $0.880$(PL) \\
        Run $3$     &   $0.507$(PI)     &   $0.933$(PL)     &   $0.667$(PI)         &   $0.933$(PL) \\
        Run $4$     &   $0.480$(PI)     &   $0.907$(BC)     &   $0.533$(PI)         &   $0.907$(BC) \\
        Run $5$     &   $0.453$(PI)     &   $0.960$(PL)     &   $0.573$(PI)         &   $0.933$(PL) \\
        Run $6$     &   $0.533$(PI)     &   $0.920$(PL)     &   $0.560$(PI)         &   $0.907$(PL) \\
        Run $7$     &   $0.547$(PI)     &   $0.960$(PL)     &   $0.560$(PI)         &   $0.947$(PL) \\
        Run $8$     &   $0.520$(PI)     &   $0.947$(PL)     &   $0.613$(PI)         &   $0.933$(PL) \\
        Run $9$     &   $0.560$(PI)     &   $0.907$(PL)     &   $0.533$(PI)         &   $0.893$(BC) \\
        Run $10$    &   $0.520$(PI)     &   $0.933$(PL)     &   $0.507$(PI)         &   $0.933$(PL) \\
        \hline
        Mean:       &   $0.509 \pm 0.031$   &   $0.931 \pm 0.026$   &   $0.565 \pm 0.048$   &   $0.919 \pm 0.020$ \\
        \hline
    \end{tabular}
    \end{center}
}
\end{table}

\begin{table}[!ht]
{\footnotesize
    \caption{Best method for \rev{the} dynamical \rev{system} dataset.}\label{tab:Dynamic_best}
    \begin{center}
    \begin{tabular}{lcll}
        \hline
        Homology            &   Accuracy    &   Vectorization           &   Classifier \\
        \hline
        $H_0$               &   $0.489$     &   Persistence Images      &   RandomForestClassifier  \\
        $H_1$               &   $0.921$     &   Persistence Landscapes  &   SVC(kernel='rbf', C=10)  \\
        $H_0+H_1$ (fused)   &   $0.553$     &   Persistence Images      &   RandomForestClassifier  \\
        $H_0+H_1$ (concat)  &   $0.905$     &   Persistence Landscapes  &   RandomForestClassifier  \\
        \hline
    \end{tabular}
    \end{center}
}
\end{table}

\subsection{MNIST}\label{4subsec2}
\begin{figure}[!ht]
     \centering
      \begin{subfigure}[b]{0.19\textwidth}
        \centering
         \includegraphics[width=0.95\textwidth]{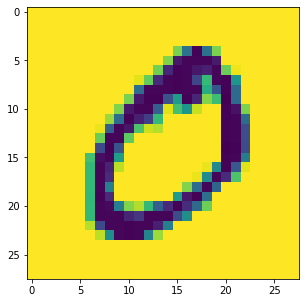}
         \caption{}
         \label{fig:MNIST:a}
     \end{subfigure}
     \begin{subfigure}[b]{0.19\textwidth}
        \centering
         \includegraphics[width=0.95\textwidth]{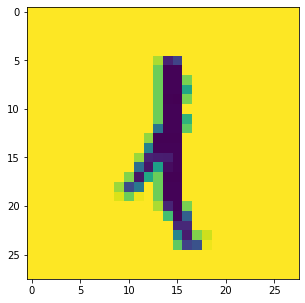}
         \caption{}
         \label{fig:MNIST:b}
     \end{subfigure}
     \begin{subfigure}[b]{0.19\textwidth}
        \centering
         \includegraphics[width=0.95\textwidth]{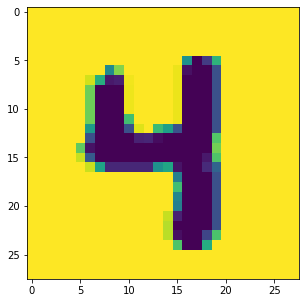}
         \caption{}
         \label{fig:MNIST:c}
     \end{subfigure}
     \begin{subfigure}[b]{0.19\textwidth}
        \centering
         \includegraphics[width=0.95\textwidth]{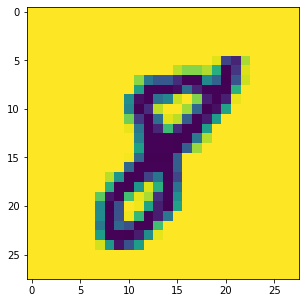}
         \caption{}
         \label{fig:MNIST:d}
     \end{subfigure}
     \begin{subfigure}[b]{0.19\textwidth}
        \centering
         \includegraphics[width=0.95\textwidth]{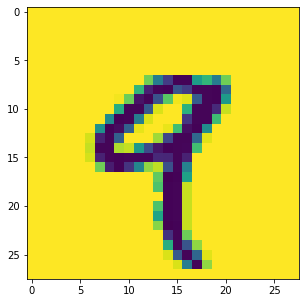}
         \caption{}
         \label{fig:MNIST:e}
     \end{subfigure}
     \caption{Sample images from MNIST dataset. It can be seen at a glance that the homology of different digits is almost always trivial or close to trivial.}
     \label{fig:MNIST}
\end{figure}

The MNIST dataset is a large dataset of handwritten digits. For more information about the MNIST Dataset, we refer the reader to \cite{deng2012mnist}, but it basically consists of $28\times 28$ pixel greyscale images representing digits, and the task is to classify each image to the corresponding digit. Figure \ref{fig:MNIST} show\rev{s}  sample images from the MNIST dataset. To limit the computational cost of this application, only a subsample composed of $5000$ training images and $1250$ test images is used, following the same train-test ratio as in Section \ref{4subsec1}. In our first, somewhat naive, approach, we directly apply the pipeline to the PDs generated by a \textit{greyscale filtration} of a normalized and negative image. The negativization of the image is needed in order to focus on the digit instead of the background. For example, in \rev{Figure \ref{fig:MNIST:d}} \del{ (middle-right)} the digit $"8"$ topologically is one connected components and two $1$-cycles. In the negative image, this is exactly what is computed by the pipeline. In the raw image instead are detected three connected components (the three yellow parts) and only one $1$-cycles. This behaviour is due to the fact that the greyscale filtration is a \textit{sublevel filtration} and so the interesting parts of the image should have low intensity. In Table \ref{tab:MNIST_runs} we report the results achieved and it is clear how they are not at all satisfactory. This behaviour is not entirely unexpected since the homology of handwritten digits is almost always trivial, with few exceptions. As discussed in Section \ref{3subsec6}, we are now going to introduce some improvements to the basic pipeline in order to achieve better results. The greyscale filtration cannot necessarily capture the difference between various digits, as their homology is similar. A switch of filtration is therefore necessary. Following \cite{8999230} we introduce the \textit{height filtration}, the \textit{radial filtration} and the \textit{density filtration}. All these filtrations capture topological features different from greyscale filtration, which simply captures the global feature of the image. We want to stress that height filtration, radial filtration and density filtration require a binarization of the image in order to be applied. The binarization of the image must be handled carefully for mainly two reasons. The first reason is the \del{inevitably} \rev{inevitable} loss of topological features during this process. The second reason is the (arbitrary) choice of the threshold, which could not be obvious a priori. Nonetheless, the MNIST dataset seems particularly suited for binarization, and a trivial threshold of $0.4$ will work just fine. We stress that, despite being referred \rev{to} as filtrations, this is in fact an abuse of notation, as they are not actually filtrations in the sense of Section \ref{3subsec2}. The following filtrations are a mere manipulation of the binarized input image, and return an image of the same dimensions. There is no persistent homology component that defines a simplicial complex to the given image. In fact, the usual cubical persistence is applied to these filtered images as well. The key difference is that the filtered images have filtration values that emphasize certain topological components.

\begin{table}[!ht]
{\footnotesize
    \caption{Accuracy for MNIST dataset.}\label{tab:MNIST_runs}
    \begin{center}
    \begin{tabular}{ccccc}
        \hline
        Accuracy:   &   $H_0$           &   $H_1$           &   $H_0+H_1$ (fused)   &   $H_0+H_1$ (concat) \\
        \hline
        Run $1$     &   $0.200$(PI)     &   $0.305$(PL)     &   $0.355$(PI)         &   $0.325$(PL) \\
        Run $2$     &   $0.177$(PI)     &   $0.318$(PL)     &   $0.346$(PI)         &   $0.319$(PL) \\
        Run $3$     &   $0.185$(PI)     &   $0.322$(PL)     &   $0.349$(PI)         &   $0.327$(PS) \\
        Run $4$     &   $0.174$(PI)     &   $0.318$(PL)     &   $0.354$(PI)         &   $0.333$(PL) \\
        Run $5$     &   $0.190$(PS)     &   $0.321$(PL)     &   $0.353$(PI)         &   $0.329$(PL) \\
        Run $6$     &   $0.182$(PI)     &   $0.326$(PL)     &   $0.356$(PI)         &   $0.338$(PL) \\
        Run $7$     &   $0.186$(PI)     &   $0.315$(PL)     &   $0.342$(PI)         &   $0.336$(PL) \\
        Run $8$     &   $0.196$(PS)     &   $0.330$(PL)     &   $0.364$(PI)         &   $0.353$(PL) \\
        Run $9$     &   $0.181$(PI)     &   $0.305$(PL)     &   $0.355$(PI)         &   $0.318$(BC) \\
        Run $10$    &   $0.182$(PI)     &   $0.318$(PL)     &   $0.355$(PI)         &   $0.325$(PL) \\
        \hline
        Mean:       &   $0.185 \pm 0.008$   &   $0.318 \pm 0.008$   &   $0.353 \pm 0.006$   &   $0.330 \pm 0.010$ \\
        \hline
    \end{tabular}
    \end{center}
}
\end{table}

\subsubsection{Height filtration}\label{4subsubsec1}
\begin{figure}[!ht]
     \begin{subfigure}[b]{0.19\textwidth}
        \centering
         \includegraphics[width=0.95\textwidth]{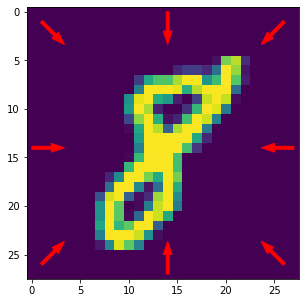}
         \caption{}
         \label{fig:MNIST_directions:a}
     \end{subfigure}
     \begin{subfigure}[b]{0.19\textwidth}
        \centering
         \includegraphics[width=0.95\textwidth]{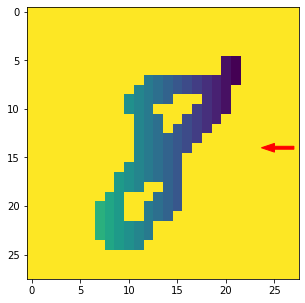}
         \caption{}
         \label{fig:MNIST_directions:b}
     \end{subfigure}
     \begin{subfigure}[b]{0.19\textwidth}
        \centering
         \includegraphics[width=0.95\textwidth]{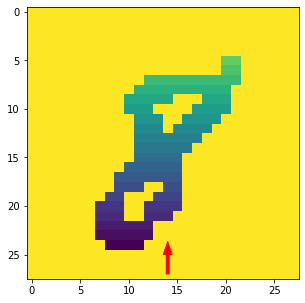}
         \caption{}
         \label{fig:MNIST_directions:c}
     \end{subfigure}
     \begin{subfigure}[b]{0.19\textwidth}
        \centering
         \includegraphics[width=0.95\textwidth]{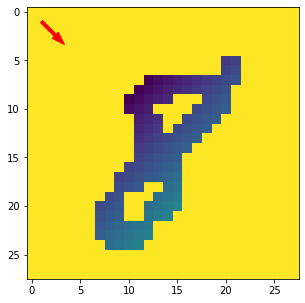}
         \caption{}
         \label{fig:MNIST_directions:d}
     \end{subfigure}
     \begin{subfigure}[b]{0.19\textwidth}
        \centering
         \includegraphics[width=0.95\textwidth]{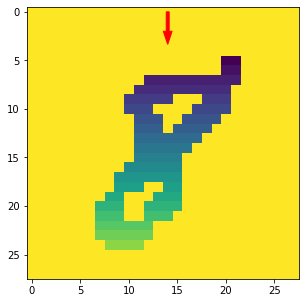}
         \caption{}
         \label{fig:MNIST_directions:e}
     \end{subfigure}
     \caption{The eight directions used for the height filtration \rev{(Figure \ref{fig:MNIST_directions:a})} and resulting filtrated images along four directions \rev{(Figures \ref{fig:MNIST_directions:b}, \ref{fig:MNIST_directions:c}, \ref{fig:MNIST_directions:d}, \ref{fig:MNIST_directions:e})}.}
     \label{fig:MNIST_directions}
\end{figure}

The height filtration detects the emergence of topological features \rev{by} looking at images only along certain directions. The birth and death value of the features is therefore linked to their position in the image, and not only to the intensity of the pixels. This is a great improvement in the case of digits, as for example along the direction $(0,1)$ the $1$-cycle of the digit $"6"$ will have a low birth value, while in the case of the digit $"9"$ the $1$-cycle will have higher birth value, and thus different PDs. With the greyscale filtration, both digits would result in one connected component and one $1$-cycle and their persistence would only be determined by the thickness of the loop. For technical reasons, linked to how the height filtration\del{'s} algorithm handles the images, it is not necessary to use the negative of the images. We now describe in more detail the height filtration, presented in \cite{turner2014persistent,8999230}. Let $\mathcal{B}\colon I\subseteq\mathbb{Z}^2 \to \{0,1\}$ be a binary image and $v \in \mathbb{R}^2$ with $\| v \|_2 = 1$. We denote with $<\cdot,\cdot>$ the Euclidean inner product. The height filtration $\mathcal{H}$ of $\mathcal{B}$ and direction $v$ is defined by
\begin{align*}
    \mathcal{H}(p):=
    \begin{cases}
        <v,p> & \text{if } \mathcal{B}(p)=1 \\
        H_\infty & \text{if } \mathcal{B}(p)=0
    \end{cases}
\end{align*}
where $p\in I$ and $H_\infty$ is the filtration value of the pixel farthest from $v$. We define eight different directions for the height filtration, namely the vectors $(0, 1)$, $(0, -1)$, $(1, 0)$, $(-1, 0)$, $(1, 1)$, $(1, -1)$, $(-1, 1)$, $(-1, -1)$. In Figure \ref{fig:MNIST_directions} we show the different directions of filtration, alongside four filtered images generated by the height filtration for different direction vectors. We highlight the fact that in \rev{Figure \ref{fig:MNIST:d}} \del{(middle-right)} and in \rev{Figure \ref{fig:MNIST_directions:a}} \del{(left)} the digit $"8"$ is one the negative of the other, for the reasons already mentioned.

\subsubsection{Radial filtration}\label{4subsubsec2}
\begin{figure}[!ht]
     \centering
     \begin{subfigure}[b]{0.19\textwidth}
        \centering
         \includegraphics[width=0.95\textwidth]{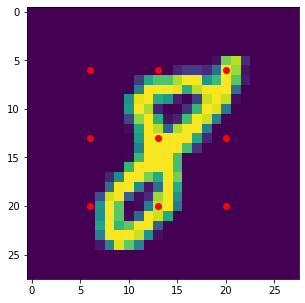}
         \caption{}
         \label{fig:MNIST_center:a}
     \end{subfigure}
     \begin{subfigure}[b]{0.19\textwidth}
        \centering
         \includegraphics[width=0.95\textwidth]{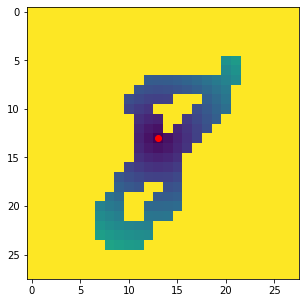}
         \caption{}
         \label{fig:MNIST_center:b}
     \end{subfigure}
     \begin{subfigure}[b]{0.19\textwidth}
        \centering
         \includegraphics[width=0.95\textwidth]{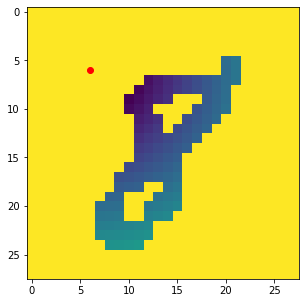}
         \caption{}
         \label{fig:MNIST_center:c}
     \end{subfigure}
     \begin{subfigure}[b]{0.19\textwidth}
        \centering
         \includegraphics[width=0.95\textwidth]{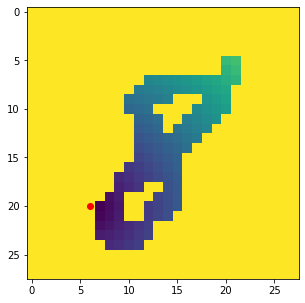}
         \caption{}
         \label{fig:MNIST_center:d}
     \end{subfigure}
     \begin{subfigure}[b]{0.19\textwidth}
        \centering
         \includegraphics[width=0.95\textwidth]{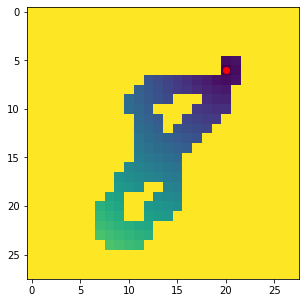}
         \caption{}
         \label{fig:MNIST_center:e}
     \end{subfigure}
     \caption{The nine centers used for the radial filtration \rev{(Figure \ref{fig:MNIST_center:a})} and resulting filtrated images with respect to four different centers \rev{(Figures \ref{fig:MNIST_center:b}, \ref{fig:MNIST_center:c}, \ref{fig:MNIST_center:d}, \ref{fig:MNIST_center:e})}.}
     \label{fig:MNIST_center}
\end{figure}

Similarly to the height filtration, the radial filtration detects the emergence of topological features as we move away from a certain center of the image. This filtration is more suited to detect heterogeneity within the image, but is way more context-dependent \del{then} \rev{than} the height filtration. In particular, the size of the image plays a crucial role in the choice of the centers. The radial filtration was introduced in \cite{atrobnm,8999230}. Given a binary image $\mathcal{B}\colon I\subseteq\mathbb{Z}^2 \to \{0,1\}$ and a center $c \in I$, the radial filtration of $\mathcal{B}$ is defined as
\begin{align*}
    \mathcal{R}(p):=
    \begin{cases}
        \|c-p\|_2 & \text{if } \mathcal{B}(p)=1 \\
        R_\infty & \text{if } \mathcal{B}(p)=0
    \end{cases}
\end{align*}
where $p \in I$ and $R_\infty$ is the filtration value of the pixel farthest from $c$. We define nine centers for the radial filtration, namely the points $(13, 6)$, $(6, 13)$, $(13, 13)$, $(20, 13)$, $(13, 20)$, $(6, 6)$, $(6, 20)$, $(20, 6)$, $(20, 20)$. In Figure \ref{fig:MNIST_center} we show the different centers of filtration, alongside four filtered images generated by the radial filtration for different centers.

\subsubsection{Density filtration}\label{4subsubsec3}
\begin{figure}[!ht]
     \centering
     \begin{subfigure}[b]{0.19\textwidth}
        \centering
         \includegraphics[width=0.95\textwidth]{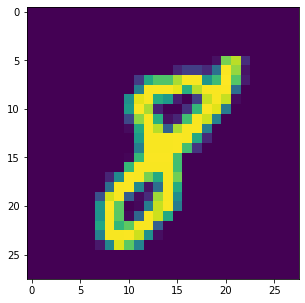}
         \caption{}
         \label{fig:MNIST_density:a}
     \end{subfigure}
     \begin{subfigure}[b]{0.19\textwidth}
        \centering
         \includegraphics[width=0.95\textwidth]{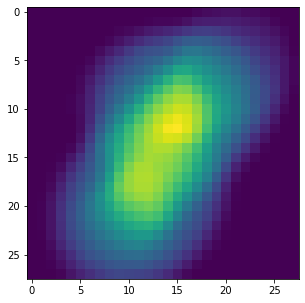}
         \caption{}
         \label{fig:MNIST_density:b}
     \end{subfigure}
     \caption{The original digit $"8"$ \rev{(Figure \ref{fig:MNIST_density:a})} and the resulting filtered image with respect to the density filtration with radius $r = 6$ \rev{(Figure \ref{fig:MNIST_density:b})}.}
     \label{fig:MNIST_density}
\end{figure}

Finally, the density filtration \cite{8999230} measures the number of lighted pixels in the neighbours of a certain pixel and is better suited to detect clusters of li\rev{gh}t\rev{ed} pixels. Given a binary image $\mathcal{B}\colon I\subseteq\mathbb{Z}^2 \to \{0,1\}$ and a radius $r \in \mathbb{R}$, the density filtration is defined as
\begin{equation*}
    \mathcal{D}(p) := \# \left\{ v \in I, \mathcal{B}(v) = 1 \text{ and } \| v - p \| \leq r \right\}
\end{equation*}
where $p \in I$ and $\| \cdot \|$ is any norm. In our pipeline, we chose $r = 6$ and the Euclidean norm. In Figure \ref{fig:MNIST_density} we show an image from the MNIST dataset and the filtered image with respect to the density filtration with $r = 6$ and the euclidean norm.

\subsubsection{Improved pipeline}\label{4subsubsec4}
We have defined eight directions for the height filtration, nine centers for the radial filtration and one radius for the density filtration. The result is that from a single image we now obtain $18$ different PDs for each homology dimension, each corresponding to a different filtration/parameter combination. Similarly to how we handled the different homology dimension\rev{s}, we follow two approaches. The first approach is to simply collapse all the PDs into one and then apply the pipeline from the PDs representation onwards. We refer to this approach as the `collapse approach'. The second idea is to first represent each one of the PDs with a vectorization method, concatenate the $18$ vectors thus resulting and use this `multivector' to classify the images. We refer to this approach as the `multivector approach'. We stress that the multivector approach and the collapse approach both result in two vectors, but the dimension of these vectors is very different. More precisely, the result of the multivector approach is $18$ times the dimension of the collapse approach. There are of course major consequences in the computational cost of the two approaches, but as this is not the purpose of our study, we will overlook these aspects. In Table \ref{tab:MNIST_collapse} we report the results of the collapse approach and in Table \ref{tab:MNIST_multivector} the results of the multivector approach. Both the approaches are very satisfactory and represent\del{s} a great improvement over the greyscale filtration. We can however notice how the representation me\rev{t}hods are not as consistent as we would like. Table \ref{tab:MNIST_best} reports the best single method for the MNIST dataset and the accuracy results are very high. This means that the non-consistency of the representation is only due to the high accuracy of every representation method, and not to a variability of the correct vectorization, thereby making the non-consistency less impactful.

\begin{table}[!ht]
{\footnotesize
    \caption{Accuracy for MNIST dataset of the collapse approach.}\label{tab:MNIST_collapse}
    \begin{center}
    \begin{tabular}{ccccc}
        \hline
        Accuracy:   &   $H_0$           &   $H_1$           &   $H_0+H_1$ (fused)   &   $H_0+H_1$ (concat) \\
        \hline
        Run $1$     &   $0.733$(PI)     &   $0.629$(PL)     &   $0.732$(PI)         &   $0.792$(PI) \\
        Run $2$     &   $0.742$(PI)     &   $0.632$(PI)     &   $0.742$(PI)         &   $0.796$(PI) \\
        Run $3$     &   $0.732$(PI)     &   $0.612$(PI)     &   $0.762$(PI)         &   $0.787$(PI) \\
        Run $4$     &   $0.739$(PI)     &   $0.639$(PL)     &   $0.753$(PI)         &   $0.806$(PI) \\
        Run $5$     &   $0.739$(PI)     &   $0.622$(PI)     &   $0.741$(PI)         &   $0.806$(PI) \\
        Run $6$     &   $0.733$(PL)     &   $0.620$(PL)     &   $0.738$(PI)         &   $0.796$(PI) \\
        Run $7$     &   $0.722$(PI)     &   $0.649$(PL)     &   $0.752$(PI)         &   $0.801$(PL) \\
        Run $8$     &   $0.716$(PI)     &   $0.635$(PL)     &   $0.738$(PI)         &   $0.779$(PI) \\
        Run $9$     &   $0.726$(PL)     &   $0.634$(PL)     &   $0.770$(PI)         &   $0.801$(PI) \\
        Run $10$    &   $0.736$(PI)     &   $0.626$(PI)     &   $0.743$(PI)         &   $0.794$(PL) \\
        \hline
        Mean:       &   $	0.732 \pm 0.008$   &   $0.630 \pm 0.010$   &   $0.747 \pm 0.011$   &   $0.796 \pm 0.008$ \\
        \hline
    \end{tabular}
    \end{center}
}
\end{table}

\begin{table}[!ht]
{\footnotesize
    \caption{Accuracy for MNIST dataset of the multivector approach.}\label{tab:MNIST_multivector}
    \begin{center}
    \begin{tabular}{ccccc}
        \hline
        Accuracy:   &   $H_0$           &   $H_1$           &   $H_0+H_1$ (fused)   &   $H_0+H_1$ (concat) \\
        \hline
        Run $1$     &   $0.911$(PL)     &   $0.614$(PL)     &   $0.944$(PI)         &   $0.937$(PL) \\
        Run $2$     &   $0.922$(PL)     &   $0.620$(PL)     &   $0.944$(BC)         &   $0.949$(PL) \\
        Run $3$     &   $0.916$(PL)     &   $0.610$(PS)     &   $0.943$(BC)         &   $0.945$(PL) \\
        Run $4$     &   $0.901$(PL)     &   $0.619$(PL)     &   $0.942$(BC)         &   $0.929$(PS) \\
        Run $5$     &   $0.911$(PL)     &   $0.601$(PL)     &   $0.937$(BC)         &   $0.942$(PL) \\
        Run $6$     &   $0.919$(PL)     &   $0.616$(PL)     &   $0.947$(BC)         &   $0.943$(PL) \\
        Run $7$     &   $0.916$(PL)     &   $0.630$(PL)     &   $0.937$(PI)         &   $0.939$(PL) \\
        Run $8$     &   $0.911$(PL)     &   $0.615$(PL)     &   $0.934$(BC)         &   $0.935$(PS) \\
        Run $9$     &   $0.918$(PL)     &   $0.617$(PL)     &   $0.946$(PL)         &   $0.944$(PL) \\
        Run $10$    &   $0.924$(PL)     &   $0.625$(PL)     &   $0.944$(BC)         &   $0.934$(PL) \\
        \hline
        Mean:       &   $	0.915 \pm 0.006$   &   $0.617 \pm 0.007$   &   $0.942 \pm 0.004$   &   $0.940 \pm 0.006$ \\
        \hline
    \end{tabular}
    \end{center}
}
\end{table}

\begin{table}[!ht]
{\footnotesize
    \caption{Best method for MNIST dataset.}\label{tab:MNIST_best}
    \begin{center}
    \begin{tabular}{lcccc}
        \hline
        Homology            &   Accuracy    &   Vectorization  &   Classifier                  &   Approach \\
        \hline
        $H_0$               &   $0.911$     &   PI             &   SVC(kernel='rbf', C=10)     &   Multivector \\
        $H_1$               &   $0.624$     &   PL             &   SVC(kernel='rbf', C=20)     &   Collapse \\
        $H_0+H_1$ (fused)   &   $0.938$     &   PI             &   SVC(kernel='rbf', C=20)     &   Multivector \\
        $H_0+H_1$ (concat)  &   $0.936$     &   PI             &   SVC(kernel='rbf', C=20)     &   Multivector \\
        \hline
    \end{tabular}
    \end{center}
}
\end{table}

\subsubsection{Comparison with other TDA approaches}\label{4subsubsec5}
Finally, a comparison is made with two papers that also use TDA and the MNIST dataset. Our improved approach is partially inspired by the work of \cite{8999230}, in which the authors perform $26$ filtrations of the same image. They compute $14$ metrics for each of these filtrations and the resulting vector of topological features for each image has a dimension of $728$ ($26$ filtrations $\times 14$ metrics $\times 2$ homology dimensions). After a feature selection using the Pearson correlation index, only $84$ not fully correlated features remain. The resulting vector is then passed to a random forest classifier. For the sake of comparison, since in our work we train only on a $5000$ images subsample of the MNIST dataset with multiple classifiers, we repeat their experiment in our context. For this reason, the results reported in this section slightly differ from those obtained in the original work. In \cite{10.3389/frai.2021.681174}, the authors perform\del{s} the height filtration along the horizontal and vertical axes, for each homology dimension, for a total of $8$ PDs. They use some vectorization methods, as well as kernel methods for PDs and an adaptive template system and then classify with the same classifiers as in our work (with the exclusion of lasso). Again, for the sake of comparison, we repeat the experiment in our context, only without the kernel method, for computational and comparison reasons. The results of both \cite{8999230} and \cite{10.3389/frai.2021.681174} are presented in Table \ref{tab:MNIST_conf}. Regarding the results obtained by \cite{10.3389/frai.2021.681174}, with (TF) we indicate that the best method was tent functions. The results obtained from our improved pipeline are in line with, if not slightly better than, those obtained from other works. It should be noted that the \del{\cite{8999230}} pipeline \rev{described in \cite{8999230}} does not use representation methods but only metrics on persistence diagrams.

\begin{table}[!ht]
{\footnotesize
    \caption{Accuracy for MNIST dataset of \cite{8999230} pipeline and \cite{10.3389/frai.2021.681174} pipeline.}\label{tab:MNIST_conf}
    \begin{center}
    \begin{tabular}{ccc}
        \hline
        Accuracy:   &   \cite{8999230} pipeline     &   \cite{10.3389/frai.2021.681174} pipeline \\
        \hline
        Run $1$     &   $0.945$     &   $0.916$(TF) \\
        Run $2$     &   $0.929$     &   $0.924$(TF) \\
        Run $3$     &   $0.934$     &   $0.926$(TF) \\
        Run $4$     &   $0.946$     &   $0.923$(PI) \\
        Run $5$     &   $0.945$     &   $0.931$(TF) \\
        Run $6$     &   $0.934$     &   $0.925$(TF) \\
        Run $7$     &   $0.956$     &   $0.926$(TF) \\
        Run $8$     &   $0.943$     &   $0.926$(PI) \\
        Run $9$     &   $0.948$     &   $0.927$(TF) \\
        Run $10$    &   $0.933$     &   $0.926$(TF) \\
        \hline
        Mean:       &   $0.941 \pm 0.008$   &   $0.925 \pm 0.004$ \\
        \hline
    \end{tabular}
    \end{center}
}
\end{table}

\subsection{FMNIST}\label{4subsec3}
\begin{figure}[!ht]
     \centering
      \begin{subfigure}[b]{0.19\textwidth}
        \centering
         \includegraphics[width=0.95\textwidth]{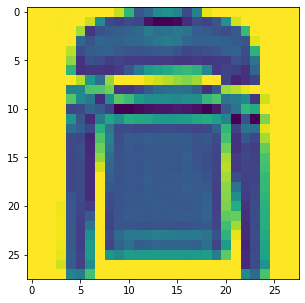}
         \caption{}
         \label{fig:FMNIST:a}
     \end{subfigure}
     \begin{subfigure}[b]{0.19\textwidth}
        \centering
         \includegraphics[width=0.95\textwidth]{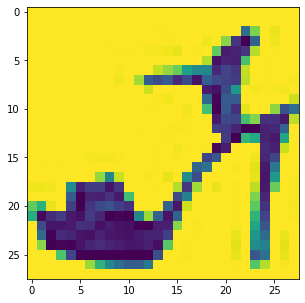}
         \caption{}
         \label{fig:FMNIST:b}
     \end{subfigure}
     \begin{subfigure}[b]{0.19\textwidth}
        \centering
         \includegraphics[width=0.95\textwidth]{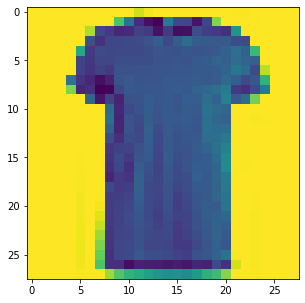}
         \caption{}
         \label{fig:FMNIST:c}
     \end{subfigure}
     \begin{subfigure}[b]{0.19\textwidth}
        \centering
         \includegraphics[width=0.95\textwidth]{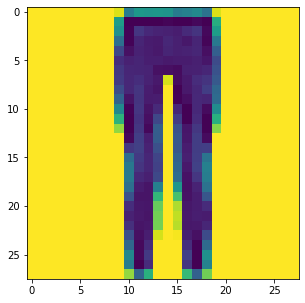}
         \caption{}
         \label{fig:FMNIST:d}
     \end{subfigure}
     \begin{subfigure}[b]{0.19\textwidth}
        \centering
         \includegraphics[width=0.95\textwidth]{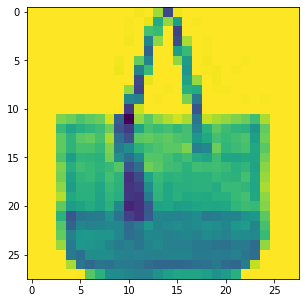}
         \caption{}
         \label{fig:FMNIST:e}
     \end{subfigure}
     \caption{Sample images from FMNIST dataset. Classifying these images is clearly more difficult than with the MNIST dataset.}
     \label{fig:FMNIST}
\end{figure}

\begin{figure}[!ht]
     \centering
     \begin{subfigure}[b]{0.19\textwidth}
        \centering
         \includegraphics[width=0.95\textwidth]{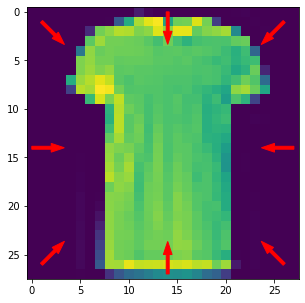}
         \caption{}
         \label{fig:FMNIST_filtrations:a}
     \end{subfigure}
     \begin{subfigure}[b]{0.19\textwidth}
        \centering
         \includegraphics[width=0.95\textwidth]{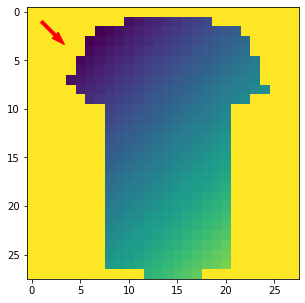}
         \caption{}
         \label{fig:FMNIST_filtrations:b}
     \end{subfigure}
     \begin{subfigure}[b]{0.19\textwidth}
        \centering
         \includegraphics[width=0.95\textwidth]{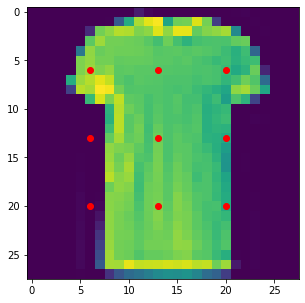}
         \caption{}
         \label{fig:FMNIST_filtrations:c}
     \end{subfigure}
     \begin{subfigure}[b]{0.19\textwidth}
        \centering
         \includegraphics[width=0.95\textwidth]{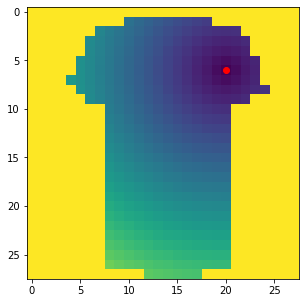}
         \caption{}
         \label{fig:FMNIST_filtrations:d}
     \end{subfigure}
     \begin{subfigure}[b]{0.19\textwidth}
        \centering
         \includegraphics[width=0.95\textwidth]{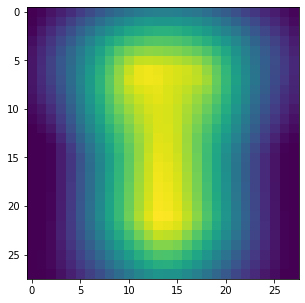}
         \caption{}
         \label{fig:FMNIST_filtrations:e}
     \end{subfigure}
     \caption{The eight directions used for the height filtration \rev{(Figure \ref{fig:FMNIST_filtrations:a})} and resulting filtrated image \rev{(Figure \ref{fig:FMNIST_filtrations:b})}. The nine centers used for the radial filtration \rev{(Figure \ref{fig:FMNIST_filtrations:c})} and \rev{the} resulting filtered image \rev{(Figure \ref{fig:FMNIST_filtrations:d})}. Density filtered image \rev{(Figure \ref{fig:FMNIST_filtrations:e})}.}
     \label{fig:FMNIST_filtrations}
\end{figure}

Fashion MNIST (FMNIST) is a dataset of Zalando's article images, consisting of $28\times 28$ pixel greyscale images divided in\rev{to} ten classes, exactly like MNIST. FMNIST is intended to be a direct drop-in replacement for the original MNIST. This dataset is strongly adequate for our study since we know that the topology of handwritten digits is almost always trivial, while this is not the case in fashion objects. For more information on the FMNIST dataset, we refer the reader to \cite{xiao2017fashionmnist}. Figure \ref{fig:FMNIST} shows sample images from the FMNIST dataset. Since the context is fairly identical to Section \ref{4subsec2}, we follow exactly the same approach. Table \ref{tab:FMNIST_runs} gives the results of the pipeline to the FMNIST dataset. Again, the results are not at all adequate, although there is a consistency in the vectorization method. Nevertheless, we can already observe a clear increase compared to the respective MNIST results (Table \ref{tab:MNIST_runs}). This is in accordance with the fact that greyscale filtration is more suitable when the homology of the image is non-trivial, as in the case of fashion images. Following the approach used in the previous section, we introduce the exact same improvements to the pipeline as the context is precisely the same. In Figure \ref{fig:FMNIST_filtrations} we show some filtered images for the FMNIST dataset. Table \ref{tab:FMNIST_collapse} and Table \ref{tab:FMNIST_multivector} report the results of respectively the collapse approach and the multivector approach for the FMNIST dataset. Both these approaches are a great improvement over the original pipeline, meaning that also for more complex images a diversification of the filtration may be well suited. The dataset is clearly more convoluted than the MNIST dataset, and this may explain the significant drop in results compared to the previous section. The consistency of the representation is however very remarkable. Following Section \ref{4subsubsec5}, we compare our results with those obtained by the \cite{8999230} and \cite{10.3389/frai.2021.681174} pipelines. We highlight the fact that in both papers the FMNIST dataset was not treated, but given the similarity of the two databases we simply reused their code for MNIST. The results of both pipelines are reported in Table \ref{tab:FMNIST_conf}. The accuracy results for this application are quite surprising. In particular, our pipeline does not perform\del{s} very well, despite the improvement from the previous section. The pipeline from \cite{8999230} achieves slightly better results, while the \cite{10.3389/frai.2021.681174} pipeline performs best of all. Possible reasons may be a different choice of parameters but also a simpler filtration which, in this particular setting, is more suited for the dataset.

\begin{table}[!ht]
{\footnotesize
    \caption{Accuracy for FMNIST dataset.}\label{tab:FMNIST_runs}
    \begin{center}
    \begin{tabular}{ccccc}
        \hline
        Accuracy:   &   $H_0$           &   $H_1$           &   $H_0+H_1$ (fused)   &   $H_0+H_1$ (concat) \\
        \hline
        Run $1$     &   $0.519$(PI)     &   $0.390$(PI)     &   $0.538$(PI)         &   $0.530$(PI) \\
        Run $2$     &   $0.499$(PI)     &   $0.398$(PI)     &   $0.544$(PI)         &   $0.524$(PI) \\
        Run $3$     &   $0.524$(PI)     &   $0.416$(PI)     &   $0.553$(PI)         &   $0.548$(PI) \\
        Run $4$     &   $0.485$(PI)     &   $0.385$(PI)     &   $0.498$(PI)         &   $0.512$(PI) \\
        Run $5$     &   $0.474$(PI)     &   $0.370$(PI)     &   $0.508$(PI)         &   $0.526$(PI) \\
        Run $6$     &   $0.491$(PI)     &   $0.381$(PI)     &   $0.511$(PI)         &   $0.516$(PI) \\
        Run $7$     &   $0.536$(PI)     &   $0.379$(PI)     &   $0.547$(PI)         &   $0.527$(PI) \\
        Run $8$     &   $0.533$(PI)     &   $0.401$(PI)     &   $0.560$(PI)         &   $0.542$(PI) \\
        Run $9$     &   $0.513$(PI)     &   $0.388$(PI)     &   $0.542$(PI)         &   $0.532$(PI) \\
        Run $10$    &   $0.495$(PI)     &   $0.373$(PI)     &   $0.523$(PI)         &   $0.528$(PI) \\
        \hline
        Mean:       &   $0.507 \pm 0.020$   &   $0.388 \pm 0.013$   &   $0.532 \pm 0.020$   &   $0.529 \pm 0.010$ \\
        \hline
    \end{tabular}
    \end{center}
}
\end{table}

\begin{table}[!ht]
{\footnotesize
    \caption{Accuracy for FMNIST dataset of the collapse approach.}\label{tab:FMNIST_collapse}
    \begin{center}
    \begin{tabular}{ccccc}
        \hline
        Accuracy:   &   $H_0$           &   $H_1$           &   $H_0+H_1$ (fused)   &   $H_0+H_1$ (concat) \\
        \hline
        Run $1$     &   $0.642$(PL)     &   $0.430$(PI)     &   $0.632$(PL)         &   $0.662$(PL) \\
        Run $2$     &   $0.627$(PL)     &   $0.410$(PI)     &   $0.611$(PL)         &   $0.640$(PL) \\
        Run $3$     &   $0.636$(PL)     &   $0.442$(PI)     &   $0.619$(PL)         &   $0.672$(PL) \\
        Run $4$     &   $0.630$(PL)     &   $0.391$(PL)     &   $0.613$(PL)         &   $0.654$(PL) \\
        Run $5$     &   $0.634$(PL)     &   $0.418$(PI)     &   $0.621$(PL)         &   $0.661$(PL) \\
        Run $6$     &   $0.620$(PL)     &   $0.410$(PI)     &   $0.612$(PL)         &   $0.642$(PL) \\
        Run $7$     &   $0.637$(PL)     &   $0.421$(PI)     &   $0.611$(PL)         &   $0.662$(PL) \\
        Run $8$     &   $0.640$(PL)     &   $0.434$(PI)     &   $0.637$(PL)         &   $0.662$(PL) \\
        Run $9$     &   $0.646$(PL)     &   $0.426$(PL)     &   $0.628$(PI)         &   $0.656$(PL) \\
        Run $10$    &   $0.627$(PL)     &   $0.413$(PL)     &   $0.610$(PL)         &   $0.646$(PL) \\
        \hline
        Mean:       &   $	0.634 \pm 0.008$   &   $0.419 \pm 0.014$   &   $0.619 \pm 0.009$   &   $0.656 \pm 0.010$ \\
        \hline
    \end{tabular}
    \end{center}
}
\end{table}

\begin{table}[!ht]
{\footnotesize
    \caption{Accuracy for FMNIST dataset of the multivector approach.}\label{tab:FMNIST_multivector}
    \begin{center}
    \begin{tabular}{ccccc}
        \hline
        Accuracy:   &   $H_0$           &   $H_1$           &   $H_0+H_1$ (fused)   &   $H_0+H_1$ (concat) \\
        \hline
        Run $1$     &   $0.678$(PL)     &   $0.431$(PL)     &   $0.750$(PL)         &   $0.717$(PL) \\
        Run $2$     &   $0.679$(PL)     &   $0.420$(PS)     &   $0.702$(PL)         &   $0.682$(PL) \\
        Run $3$     &   $0.704$(PL)     &   $0.448$(PL)     &   $0.718$(PL)         &   $0.715$(PL) \\
        Run $4$     &   $0.690$(PL)     &   $0.408$(PL)     &   $0.721$(PL)         &   $0.706$(PL) \\
        Run $5$     &   $0.678$(PL)     &   $0.418$(PL)     &   $0.714$(PL)         &   $0.707$(PL) \\
        Run $6$     &   $0.670$(PL)     &   $0.397$(PL)     &   $0.707$(PL)         &   $0.678$(PL) \\
        Run $7$     &   $0.686$(PL)     &   $0.412$(PL)     &   $0.705$(PL)         &   $0.688$(PL) \\
        Run $8$     &   $0.698$(PL)     &   $0.425$(PL)     &   $0.721$(PL)         &   $0.712$(PL) \\
        Run $9$     &   $0.690$(PL)     &   $0.438$(PL)     &   $0.716$(PL)         &   $0.707$(PL) \\
        Run $10$    &   $0.682$(PL)     &   $0.414$(PL)     &   $0.709$(PL)         &   $0.696$(PL) \\
        \hline
        Mean:       &   $	0.686 \pm 0.010$   &   $0.421 \pm 0.014$   &   $0.716 \pm 0.013$   &   $0.701 \pm 0.013$ \\
        \hline
    \end{tabular}
    \end{center}
}
\end{table}

\begin{table}[!ht]
{\footnotesize
    \caption{Best method for FMNIST dataset.}\label{tab:FMNIST_best}
    \begin{center}
    \begin{tabular}{lcccc}
        \hline
        Homology            &   Accuracy    &   Vectorization  &   Classifier                  &   Approach \\
        \hline
        $H_0$               &   $0.681$     &   PL             &   RFC     &   Multivector \\
        $H_1$               &   $0.417$     &   PI             &   RFC     &   Collapse \\
        $H_0+H_1$ (fused)   &   $0.716$     &   PL             &   RFC     &   Multivector \\
        $H_0+H_1$ (concat)  &   $0.701$     &   PL             &   RFC     &   Multivector \\
        \hline
    \end{tabular}
    \end{center}
}
\end{table}

\begin{table}[!ht]
{\footnotesize
    \caption{Accuracy for FMNIST dataset of \cite{8999230} pipeline and \cite{10.3389/frai.2021.681174} pipeline.}\label{tab:FMNIST_conf}
    \begin{center}
    \begin{tabular}{ccc}
        \hline
        Accuracy:   &   \cite{8999230} pipeline     &   \cite{10.3389/frai.2021.681174} pipeline \\
        \hline
        Run $1$     &   $0.753$     &   $0.810$(PI) \\
        Run $2$     &   $0.739$     &   $0.795$(PI) \\
        Run $3$     &   $0.750$     &   $0.818$(TF) \\
        Run $4$     &   $0.757$     &   $0.793$(PI) \\
        Run $5$     &   $0.769$     &   $0.795$(PI) \\
        Run $6$     &   $0.738$     &   $0.792$(PI) \\
        Run $7$     &   $0.750$     &   $0.802$(PI) \\
        Run $8$     &   $0.748$     &   $0.813$(PI) \\
        Run $9$     &   $0.752$     &   $0.803$(PI) \\
        Run $10$    &   $0.736$     &   $0.815$(PI) \\
        \hline
        Mean:       &   $0.749 \pm 0.009$   &   $0.804 \pm 0.009$ \\
        \hline
    \end{tabular}
    \end{center}
}
\end{table}

\subsection{COLLAB}\label{4subsec4}
\begin{figure}[!ht]
     \centering
       \begin{subfigure}[b]{0.45\textwidth}
        \centering
         \includegraphics[width=0.95\textwidth]{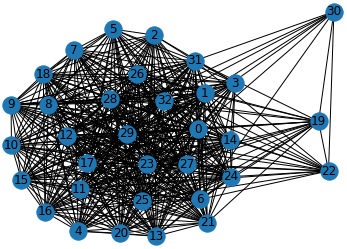}
         \caption{}
         \label{fig:COLLAB:a}
     \end{subfigure}
     \begin{subfigure}[b]{0.45\textwidth}
        \centering
         \includegraphics[width=0.95\textwidth]{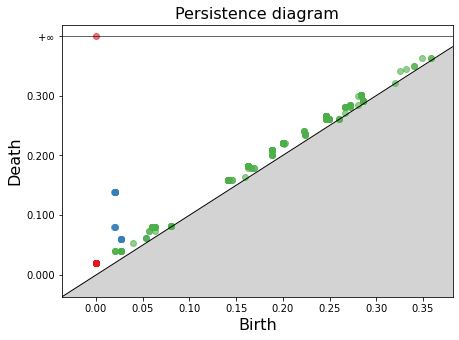}
         \caption{}
         \label{fig:COLLAB:b}
     \end{subfigure}
     \caption{A graph of COLLAB \rev{(Figure \ref{fig:COLLAB:a})} and the corresponding PD \rev{(Figure \ref{fig:COLLAB:b})}. For aesthetic reasons, only a small sample of $2$-simplexes (green points) are shown and the edge weight is not displayed.}
     \label{fig:COLLAB}
\end{figure}

COLLAB dataset is a network graphs dataset of scientific collaborations coming from \cite{10.1145/2783258.2783417}. It consists of $5000$ graphs derived from three public collaboration
datasets which also serve as labels: \textit{high energy physics}, \textit{condensed matter physics} and \textit{astro physics}. Each node of the graphs is an author, and there is a link between two authors if they coauthor a scientific article. COLLAB is a dataset of weighted, undirected graphs. Every collaboration between $n$ authors contribute\rev{s} to the edge weight between those authors of a factor $1/(n-1)$. The vertices are not weighted, this means that all vertices immediately enter the filtration as $0$-simplexes. The filtration value of the $1$-simplexes is the weight of the edge connecting the two vertices and for $2$-simplexes we chose as filtration value the maximum weight of the edges forming it. For computational reasons, the maximum homology dimension computed is $H_2$. In Figure \ref{fig:COLLAB} we show a graph of COLLAB and the associated PD. For aesthetic reasons we have included only a small portion of the $2$-simplexes and the edge weight is not displayed. The graphs of this dataset are extremely connected and the computational cost in order to compute their PD is very high. In Table \ref{tab:COLLAB} we report the results achieved by the pipeline on COLLAB dataset, and in Table \ref{tab:COLLAB_best} the best combination representation - classifier. These results are quite satisfactory and in line with other topology-based methods, such as PersLay \cite{CarriereCILRU20}, achieving an accuracy of $76.4\%$ (mean accuracy over ten runs of a 10-fold classification evaluation).

\begin{table}[!ht]
{\footnotesize
    \caption{Accuracy for COLLAB dataset.}\label{tab:COLLAB}
    \begin{center}
    \begin{tabular}{cccccc}
        \hline
        Accuracy:   &   $H_0$           &   $H_1$           &   $H_2$   &   $H_0+H_1+H_2$ (fused)   &   $H_0+H_1+H_2$ (concat) \\
        \hline
        Run $1$     &   $0.602$(PI)     &   $0.549$(PS)     &   $0.731$(PI)   &   $0.730$(PI)         &   $0.730$(PI) \\
        Run $2$     &   $0.613$(PI)     &   $0.543$(PS)     &   $0.760$(PI)   &   $0.759$(PI)         &   $0.747$(PI) \\
        Run $3$     &   $0.613$(PI)     &   $0.549$(PS)     &   $0.741$(PI)   &   $0.747$(PI)         &   $0.739$(PI) \\
        Run $4$     &   $0.621$(BC)     &   $0.542$(PL)     &   $0.736$(PI)   &   $0.749$(PI)         &   $0.737$(PI) \\
        Run $5$     &   $0.628$(BC)     &   $0.551$(PS)     &   $0.746$(PI)   &   $0.758$(PI)         &   $0.752$(PI) \\
        Run $6$     &   $0.621$(PI)     &   $0.557$(PS)     &   $0.759$(PI)   &   $0.763$(PI)         &   $0.753$(PI) \\
        Run $7$     &   $0.609$(PI)     &   $0.550$(PS)     &   $0.736$(PI)   &   $0.750$(PI)         &   $0.734$(PI) \\
        Run $8$     &   $0.626$(BC)     &   $0.557$(PS)     &   $0.750$(PI)   &   $0.751$(PI)         &   $0.725$(PI) \\
        Run $9$     &   $0.615$(PS)     &   $0.559$(PS)     &   $0.745$(PI)   &   $0.749$(PI)         &   $0.739$(PI) \\
        Run $10$    &   $0.607$(PS)     &   $0.567$(PS)     &   $0.753$(PI)   &   $0.748$(PI)         &   $0.739$(PI) \\
        \hline
        Mean:       &   $	0.616 \pm 0.008$   &   $0.552 \pm 0.007$   &   $0.746 \pm 0.009$&   $0.750 \pm 0.009$   &   $0.739 \pm 0.009$ \\
        \hline
    \end{tabular}
    \end{center}
}
\end{table}

\begin{table}[!ht]
{\footnotesize
    \caption{Best method for COLLAB dataset.}\label{tab:COLLAB_best}
    \begin{center}
    \begin{tabular}{lcll}
        \hline
        Homology            &   Accuracy    &   Vectorization           &   Classifier \\
        \hline
        $H_0$               &   $0.613$     &   Betti Curves      &   RandomForestClassifier  \\
        $H_1$               &   $0.550$     &   Persistence Silhouette  &   RandomForestClassifier  \\
        $H_2$               &   $0.746$     &   Persistence Images  &   RandomForestClassifier  \\
        $H_0+H_1+H_2$ (fused)   &   $0.749$     &   Persistence Images      &   RandomForestClassifier  \\
        $H_0+H_1+H_2$ (concat)  &   $0.736$     &   Persistence Images  &   RandomForestClassifier  \\
        \hline
    \end{tabular}
    \end{center}
}
\end{table}
\section{Discussion}\label{sec5}
The proposed pipeline proved to be a valuable classification tool in various contexts. Moreover, some patterns emerged in the course of the experiments. In particular, for point clouds and graphs, it seems that the maximum homology dimension alone is sufficient to obtain very appreciable results. For images, on the other hand, only $H_0$ achieves good results. The `fused' and `concat' approaches are consistently among the best performers, with the exception of the dynamic dataset where `fused' fails. This discrepancy does not seem to be explained by heterogeneity in the number of points in the homology dimensions, which are comparable in all datasets with the exception of COLLAB. For these reasons, it would seem that regardless the type of data under consideration and the number of points in the different homology dimensions, the `concat' approach is the safest method. Alternatively, if the data are not images and one wishes to reduce the computational cost, using only the maximum homology dimension seems to be a viable option. No correlation emerged between data type and vectorization method. In general, it seems that persistence image and persistence landscape are always the safest options. 
\rev{In order to support these statements, an analysis of the statistical significance has been performed (using the paired t-test) on a subset of the results presented in Section \ref{sec4}. In the following, each table shows the p-values associated with the mean accuracy results of the different vectorization methods for each homology dimension. For the sake of brevity, we describe in detail only Table \ref{tab:Dynamic_statistic} and Table \ref{tab:Homology_statistic}. Table \ref{tab:Dynamic_statistic} has been computed as follows. In Section \ref{sec4} we have computed the mean accuracy of each classifier and each vectorization method previously described over the course of the ten runs of the cross-validation. This consists of a matrix of $9$ rows (one for each classifier) and $21$ columns (one for each vectorization method). We have selected the best row (in terms of mean accuracy) for each vectorization method and tested each vectorization method against each other using the t-test function from scipy. For more information on t-test and scipy we refer the reader to \cite{kim2015t,2020SciPy-NMeth}. In Section \ref{sec4} we have assessed that PI is the best vectorization method for $H_0$ and $H_0 + H_1$ `fused', while PL is the best method for $H_1$ and $H_0 + H_1$ `concat'. Table \ref{tab:Dynamic_statistic} ensures that these statements have statistical validity, because the p-value associated with these methods and homology dimensions is sufficiently small. In Table \ref{tab:Homology_statistic} we have followed the same procedure, with the difference that we have compared each homological dimension against each other for every dataset. In particular, the p-value of the `concat' approach is consistently small, confirming that its usefulness is supported by statistical results.}

\begin{table}[H] 
\caption{\rev{p-value statistic for the dynamical system dataset.}\label{tab:Dynamic_statistic}}
\newcolumntype{C}{>{\centering\arraybackslash}X}
\begin{tabularx}{\textwidth}{C|CCCC}
\toprule
\textbf{p-value}  &   \textbf{$H_0$}	& \textbf{$H_1$}	& \textbf{$H_0+H1$} fused	& \textbf{$H_0+H1$} concat\\
\midrule
\textbf{PI vs PL}		& $2.03\cdot10^{-9}$			& $9.33\cdot10^{-3}$			& $1.27\cdot10^{-3}$			& $5.06\cdot10^{-2}$\\
\textbf{PI vs PS}		& $2.03\cdot10^{-9}$			& $3.44\cdot10^{-1}$			& $5.38\cdot10^{-1}$			& $7.97\cdot10^{-1}$\\
\textbf{PI vs BC}		& $2.03\cdot10^{-9}$			& $4.60\cdot10^{-3}$			& $1.27\cdot10^{-3}$			& $3.84\cdot10^{-3}$\\
\textbf{PL vs PS}		& Null			& $1.28\cdot10^{-3}$			& $2.26\cdot10^{-4}$			& $1.34\cdot10^{-4}$\\
\textbf{PL vs BC}		& Null			& $4.61\cdot10^{-2}$			& Null			& $2.55\cdot10^{-3}$\\
\textbf{PS vs BC}		& Null			& $7.90\cdot10^{-3}$			& $2.26\cdot10^{-4}$			& $1.37\cdot10^{-4}$\\
\bottomrule
\end{tabularx}
\end{table}

\begin{table}[H] 
\caption{\rev{p-value statistic for the MNIST dataset.}\label{tab:MNIST_statistic}}
\newcolumntype{C}{>{\centering\arraybackslash}X}
\begin{tabularx}{\textwidth}{C|CCCC}
\toprule
\textbf{p-value}  &   \textbf{$H_0$}	& \textbf{$H_1$}	& \textbf{$H_0+H1$} fused	& \textbf{$H_0+H1$} concat\\
\midrule
\textbf{PI vs PL}		& $4.55\cdot10^{-9}$			& $1.02\cdot10^{-8}$			& $1.82\cdot10^{-2}$			& $3.39\cdot10^{-7}$\\
\textbf{PI vs PS}		& $8.20\cdot10^{-7}$			& $6.32\cdot10^{-7}$			& $1.58\cdot10^{-2}$			& $3.98\cdot10^{-6}$\\
\textbf{PI vs BC}		& $1.21\cdot10^{-3}$			& $8.97\cdot10^{-9}$			& $2.73\cdot10^{-1}$			& $4.18\cdot10^{-5}$\\
\textbf{PL vs PS}		& $1.06\cdot10^{-4}$			& $2.27\cdot10^{-2}$			& $6.23\cdot10^{-2}$			& $9.62\cdot10^{-5}$\\
\textbf{PL vs BC}		& $1.14\cdot10^{-6}$			& $3.62\cdot10^{-2}$			& $4.27\cdot10^{-2}$			& $6.76\cdot10^{-5}$\\
\textbf{PS vs BC}		& $7.92\cdot10^{-5}$			& $4.70\cdot10^{-3}$			& $4.91\cdot10^{-3}$			& $1.46\cdot10^{-3}$\\
\bottomrule
\end{tabularx}
\end{table}

\begin{table}[H] 
\caption{\rev{p-value statistic for the FMNIST dataset.\label{tab:FMNIST_statistic}}}
\newcolumntype{C}{>{\centering\arraybackslash}X}
\begin{tabularx}{\textwidth}{C|CCCC}
\toprule
\textbf{p-value}  &   \textbf{$H_0$}	& \textbf{$H_1$}	& \textbf{$H_0+H1$} fused	& \textbf{$H_0+H1$} concat\\
\midrule
\textbf{PI vs PL}		& $2.02\cdot10^{-2}$			& $6.40\cdot10^{-5}$			& $2.77\cdot10^{-1}$			& $9.63\cdot10^{-3}$\\
\textbf{PI vs PS}		& $1.34\cdot10^{-1}$			& $2.54\cdot10^{-5}$			& $1.01\cdot10^{-3}$			& $6.45\cdot10^{-3}$\\
\textbf{PI vs BC}		& $3.96\cdot10^{-1}$			& $1.96\cdot10^{-3}$			& $2.68\cdot10^{-2}$			& $8.23\cdot10^{-1}$\\
\textbf{PL vs PS}		& $6.91\cdot10^{-3}$			& $7.16\cdot10^{-1}$			& $9.79\cdot10^{-1}$			& $2.67\cdot10^{-1}$\\
\textbf{PL vs BC}		& $1.53\cdot10^{-2}$			& $2.05\cdot10^{-6}$			& $3.97\cdot10^{-3}$			& $1.66\cdot10^{-3}$\\
\textbf{PS vs BC}		& $6.25\cdot10^{-1}$			& $1.04\cdot10^{-6}$			& $8.16\cdot10^{-3}$			& $3.46\cdot10^{-4}$\\
\bottomrule
\end{tabularx}
\end{table}

\begin{table}[H] 
\caption{\rev{p-value statistic for the COLLAB dataset.}\label{tab:COLLAB_statistic}}
\newcolumntype{C}{>{\centering\arraybackslash}X}
\begin{tabularx}{\textwidth}{C|CCCCC}
\toprule
\textbf{p-value}  &   \textbf{$H_0$}	& \textbf{$H_1$}	& \textbf{$H_2$}	& \textbf{$H_0+H1$} fused	& \textbf{$H_0+H1$} concat\\
\midrule
\textbf{PI vs PL}		& $3.73\cdot10^{-2}$			& $1.53\cdot10^{-5}$			& $7.87\cdot10^{-4}$			& $7.89\cdot10^{-4}$		& $1.13\cdot10^{-3}$\\
\textbf{PI vs PS}		& $1.52\cdot10^{-1}$			& $7.06\cdot10^{-5}$			& $5.17\cdot10^{-3}$			& $2.32\cdot10^{-4}$		& $7.29\cdot10^{-4}$\\
\textbf{PI vs BC}		& $7.80\cdot10^{-1}$			& $3.71\cdot10^{-4}$			& $1.13\cdot10^{-3}$			& $2.01\cdot10^{-5}$		& $8.58\cdot10^{-5}$\\
\textbf{PL vs PS}		& $3.14\cdot10^{-4}$			& $4.24\cdot10^{-1}$			& $3.36\cdot10^{-5}$			& $2.67\cdot10^{-1}$		& $2.56\cdot10^{-1}$\\
\textbf{PL vs BC}		& $4.45\cdot10^{-5}$			& $9.19\cdot10^{-3}$			& $6.17\cdot10^{-2}$			& $2.20\cdot10^{-1}$		& $3.46\cdot10^{-1}$\\
\textbf{PS vs BC}		& $7.63\cdot10^{-4}$			& $7.05\cdot10^{-3}$			& $8.99\cdot10^{-1}$			& $7.00\cdot10^{-1}$		& $3.31\cdot10^{-2}$\\
\bottomrule
\end{tabularx}
\end{table}

\begin{table}[H] 
\caption{\rev{p-value statistic for the homology dimensions.}\label{tab:Homology_statistic}}
\newcolumntype{C}{>{\centering\arraybackslash}X}
\begin{tabularx}{\textwidth}{C|CCCC}
\toprule
\textbf{p-value}  &   \textbf{Dynamical}	& \textbf{MNIST}	& \textbf{FMNIST}	& \textbf{COLLAB}\\
\midrule
$H_0$ vs $H_1$		& $5.71\cdot10^{-5}$			& $1.82\cdot10^{-10}$			& $1.78\cdot10^{-8}$			& $1.39\cdot10^{-10}$\\
$H_0$ vs $H_2$		& -			& -			& -			& $9.82\cdot10^{-3}$\\
$H_0$ vs fused		& $1.35\cdot10^{-1}$			& $7.81\cdot10^{-6}$			& $5.26\cdot10^{-4}$			& $1.31\cdot10^{-2}$\\
$H_0$ vs concat		& $2.18\cdot10^{-6}$			& $3.69\cdot10^{-3}$			& $3.58\cdot10^{-2}$			& $3.87\cdot10^{-3}$\\
$H_1$ vs $H_2$		& -			& -			& -			& $3.84\cdot10^{-5}$\\
$H_1$ vs fused		& $3.56\cdot10^{-4}$			& $5.27\cdot10^{-10}$			& $1.01\cdot10^{-8}$			& $3.65\cdot10^{-5}$\\
$H_1$ vs concat		& $2.82\cdot10^{-2}$			& $5.66\cdot10^{-14}$			& $3.82\cdot10^{-8}$			& $1.34\cdot10^{-5}$\\
$H_2$ vs fused		& -			& -			& -			& $5.73\cdot10^{-1}$\\
$H_2$ vs concat		& -			& -			& -			& $3.50\cdot10^{-1}$\\
fused vs concat		& $8.08\cdot10^{-6}$			& $1.31\cdot10^{-1}$			& $1.04\cdot10^{-4}$			& $1.15\cdot10^{-2}$\\
\bottomrule
\end{tabularx}
\end{table}

\section{Conclusion}\label{sec6}
Results of classification discussed in the previous sections show that the proposed pipeline is a procedure able to maximize the capabilities of topological data analysis and machine learning. Such pipeline allows the analysis of digital data without restrictions such as data type or acquisition method. Moreover, the pipeline is not limited to the size of the dataset, which is often the case of \rev{the} most recent and best\rev{-}performing method\rev{s} for classification based on deep learning architectures. In addition, interesting correlations arose between homology dimension and classification results. The concatenation of PDs in the different homology dimensions consistently seems to be the most suitable choice. In the very near future\rev{,} we will \del{show the pipeline's classification capabilities in very challenging settings} \rev{further investigate the correlation between filtration, vectorization and data type in very challenging settings arising from real-world datasets}; e.g. remote sensing (for climate prediction), and in Raman spectroscopy (for cancer staging).

%
\bibliographystyle{splncs04}
\bibliography{References_TDA}

\end{document}